\documentclass[10pt,twocolumn,twoside]{IEEEtran}

\usepackage{graphicx}
\usepackage{caption}
\usepackage{array}
\usepackage{subfig}
\usepackage{cite}
\usepackage{stmaryrd}
\usepackage{amsfonts,latexsym,amssymb}
\usepackage[cmex10]{amsmath}
\usepackage{algorithm}
\usepackage{algorithmic}
\usepackage{hyperref}
\usepackage{multirow}
\usepackage{arydshln}

\usepackage[OT1]{fontenc}

\newtheorem{lemma}{{Lemma}}

\newtheorem{proposition}{{Proposition}}

\input{zgx.math}

\begin{document}

%
\title{Learning the Hierarchical Parts of Objects by Deep Non-Smooth Nonnegative Matrix Factorization}

\author{Jinshi Yu, Guoxu Zhou, Andrzej~Cichocki ~\IEEEmembership{IEEE Fellow},  and Shengli~Xie \IEEEmembership{IEEE Senior Member}

\thanks{J. Yu, G. Zhou, and S. Xie are with the Faculty of Automation, Guangdong University of
Technology, Guangzhou, 510006, China
(E-mail: jinshi.yu@foxmail.com, gx.zhou@gdut.edu.cn, shlxie@gdut.edu.cn).}
\thanks{A. Cichochi is from RIKEN, Japan and SKOLTECH, Moscow, Russia}}


\maketitle


\begin{abstract}
Nonsmooth Nonnegative Matrix Factorization (\emph{ns}NMF) is capable of producing more localized, less overlapped feature representations than other variants of NMF while keeping satisfactory fit to  data. However, nsNMF as well as other existing NMF methods  is incompetent to learn hierarchical features of complex data due to its shallow structure. To fill this gap, we propose a deep \emph{ns}NMF method coined by the fact that it possesses a deeper architecture compared with standard \emph{ns}NMF. The deep \emph{ns}NMF not only gives parts-based features due to the nonnegativity constraints, but also creates higher-level, more abstract  features by combing lower-level ones. The in-depth description of how deep architecture can help to efficiently discover abstract features in dnsNMF is presented. And we also show that  the deep \emph{ns}NMF has close relationship with the deep autoencoder, suggesting that the proposed model inherits the major advantages from both deep learning and NMF. Extensive experiments demonstrate the standout performance of the proposed method in clustering analysis.

\end{abstract}


%
\IEEEpeerreviewmaketitle


\section{Introduction}
\label{section_1}
\IEEEPARstart{N}{onnegative} matrix factorization (NMF) is a technique that represents a nonnegative data matrix $\bf X$ as the product of two nonnegative matrices $\bf Z$ and $\bf H$, i.e., ${\bf X}=\bf ZH$, where $\bf Z$ and $\bf H$ are often called features and feature representation, respectively. Due to the nonnegativity constraints on $\bf Z$ and $\bf H$, the representation is purely based on additive (without subtractive) combinations of entries. As a result, the columns of $\bf Z$ can be viewed as basic building-blocks that compose $\bf X$, and the representation matrix \mat{H}, often very sparse, explains how to construct \mat{X} by superposing some of these building-blocks.  For this reason, NMF has been a powerful tool in nonnegative data analysis and successfully applied in diverse areas
     including, but not limited to, environmetrics \cite{paatero1994positive}, microarray data
     analysis \cite{brunet2004metagenes}, \cite{devarajan2008nonnegative}, document clustering \cite{xu2003document}, \cite{shahnaz2006document}, \cite{berry2005email}, face and objects
     recognition \cite{lee1999learning}, \cite{zafeiriou2006exploiting}, \cite{kotsia2007novel}, \cite{li2001learning}, \cite{guillamet2002non}, \cite{ramanath2003eigenviews}, blind audio source
     separation \cite{weninger2012optimization}, biomedical engineering \cite{kim2003subsystem}, \cite{heger2003sensitive}, \cite{paatero2003advanced}, \cite{brunet2004metagenes},
     polyphonic music transcription \cite{smaragdis2003non} and more.

Nevertheless,  NMF often produces features with a very high degree of overlapping in practice, which contradicts our intuition of ``parts-based'' learning ability of NMF.
In order to extract more localized, meaningful, and less overlapped features of the data, quite a number of NMF variants have been proposed, including local nonnegative matrix factorization \cite{feng2002local}, nonnegative sparse coding \cite{hoyer2002non}, NMF with sparseness constraints \cite{hoyer2004non,liu2003non}, just to mention a few. These NMF variants enforce sparseness of the features $\bf Z$ or the feature representation $\bf H$, or both by using norm penalty. Unfortunately, using this method to improve the sparseness of either one of $\bf Z$ and $\bf H$ will eventually reduce the sparseness of the other one in order to keep reasonable fitting to the data matrix \mat{X}. And improving the sparseness of both will certainly increase the constructed error between $\bf X$ and $\bf ZH$. To solve this dilemma, Pascual-Montano et al.  proposed a non-smooth NMF (\emph{ns}NMF) method by introducing a ``smoothing'' matrix \mat{S} into the NMF model \cite{pascual2006nonsmooth}, such that ${\bf X}=\bf ZSH$. The distinct advantage of \emph{ns}NMF is that, by adjusting the smoothing factor $\bf S$, the sparsity of both the features matrix $\bf Z$ and feature representation $\bf H$ can be finely tuned while still keeping  faithful fit of model to the original data. This feature makes \emph{ns}NMF quite flexible and versatile  in practical applications.

Usually, we perceive and understand a new object using concepts we have already known. Consequently, we recognize the world essentially using a hierarchical structure which presents many levels of understanding and abstraction of objects. In the lowest level, very detailed and concrete physical features are presented, whereas in the highest level, details are largely ignored and more abstract features are formed by combing lower level features. In other words, each level in this hierarchical structure creates a specific level of abstract perception. This has motivated the rapid development of deep learning, a powerful technique capable of  extracting high-level abstract features of objects and  has achieved great success in a wide range of machine learning tasks.  As \emph{ns}NMF always tends to learn low level features such as contours, corner, angles, and surface boundaries due to its shallow structure and sparseness enforcement, it is therefore meaningful to equip \emph{ns}NMF with a deep architecture which allows it to extract parts-based and more abstract features of objects. To serve this purpose, we develop a new model called Deep \emph{ns}NMF (dnsNMF) in this paper to perform deep decompositions in the sense that it contains multiple layers of \emph{ns}NMF. The deep architecture of dnsNMF is expected to learn high level, more abstract features by combining lower level features, in order to model more complex data.

Note that Cichocki and Zdunek have proposed a simple multilayer NMF \cite{cicho2007multi,cichocki2007hierarchical,cichocki2006multi} to improve the performance of blind signal separation (BSS) about ten years ago. Although it is a good idea at that time, their works did not consider the solution to update the parameters globally and not the mechanism to keep the good fit of model to the original data. As the success of deep learning, deep learning methods have been widely use in NMF field, such as N-NMF \cite{zhang2016nonlinear}, deep semi-NMF \cite{trigeorgis2017deep} and SDNMF \cite{guo2017sparse}. The key idea of \cite{zhang2016nonlinear} is performing the NMF/GNMF on the nonlinear dimension reduction of original data by resort to the deep learning method. Obviously, this method does not use the multilayer NMF and may lose some important information of raw data for the NMF task. In \cite{trigeorgis2017deep}, a complete deep architecture was introduced to build a multi-layer semi-NMF and also consider many other nonlinear activation on factor matrices. Unfortunately deep semi-NMF ignores nonnegative property of the learned features and the sparse structure hidden in complex data. Recently, \cite{guo2017sparse} adds the sparse constraints to factor matrix features or discriminative representation in the multilayer NMF and successfully explores the sparse structure of data by using the Nesterov's accelerated gradient descent algorithm. However, using this method to improve the sparseness of some factor matrices will reduce the sparseness of other ones in order to keep reasonable fitting to original data. Moveover, although these deep NMF methods achieve much improvement in complex data analysis, as we all know, they all lack of many important and necessary discussion, such as how a deep structure can help to efficiently discover abstract features in deep NMF and the relationship between deep NMF and deep learning. All these aspects have been considered in our paper. In particularly, we build dnsNMF model by stacking one-layer \emph{ns}NMF into multiple layers, as \emph{ns}NMF has the advantage of simultaneously improving the sparseness of all factor matrices while keeping faithful fit of model to the original data. The reason that how a deep NMF can efficiently discover abstract features is described in \ref{abstract feature description}. And the description of relationship between deep NMF and deep learning is presented in Section 4.

The rest of the paper is organized as follows: In Section 2, a brief review of NMF  and \emph{ns}NMF is presented. In Section 3, the dnsNMF method is proposed and implementation details are provided.
 The relationship between dnsNMF and deep autoencoding is briefly explained in Section 4. How dnsNMF can be used to extract a hierarchy of different level of abstract features is  visualized and experimental results on clustering are presented in Section 5. Finally, some concluding remarks and future works are given in Section 6.

\section{Background}
\label{section_2}
\subsection{Nonnegative Matrix Factorization (NMF)}
We assume that the data matrix  ${\bf X}=\lbrack {\boldsymbol x}_{1},{\boldsymbol x}_{2},...{\boldsymbol x}_{n}\rbrack\in{\bf \Re}^{p\times n}_{+}$ is  a collection of $n$ data samples as its columns, each with $p$ features. Nonnegative matrix factorization (NMF) \cite{lee2001algorithms} finds a compact representation of $\bf X$ such that
\begin{equation}\label{eq_1}
\bf{X}\approx\bf{ZH}, \; \bf{Z}\ge\bf{0}, \bf{H}\ge \bf{0},
\end{equation}
where ${\bf Z}\in{\bf \Re}^{p\times r}_{+} $ is the matrix of basis vectors, and ${\bf H}\in{\bf\Re}^{r\times n}_{+}$ the encoding vectors or projections, $r\le \min\set{p,n}$, all matrices $\bf X$, $\bf Z$, $\bf H$ are nonnegative.

In practice, NMF can be implemented by minimizing the distance between the original data $\bf X$ and its low-rank representation  $\bf{ZH}$. One of the most widely used distance measurements is the Euclidean distance, defined as:
\begin{equation}\label{eq_2}
 E\left({\bf Z}, {\bf H}\right)=\frac{1}{2}\sum\limits_{i=1}^{p}\sum\limits_{j=1}^{n}\left({\bf X}_{ij}-{\left(\bf ZH\right)}_{ij}\right)^{2}=\frac{1}{2}{\|\bf X-\bf ZH\|}_{F}^{2}.
\end{equation}
Lee and Seung derived the following multiplicative update rules \cite{lee2001algorithms} to minimize the cost function \eqref{eq_2}:
\\
\begin{equation}\label{eq:MU_H}
{\bf H}\leftarrow{\bf H}\hdp\frac{{\bf Z}^{T}{\bf X}}{{\bf Z}^{T}{\bf ZH}},
\end{equation}
\begin{equation}\label{eq:MU_Z}
{\bf Z}\leftarrow{\bf Z}\hdp\frac{{\bf X}{\bf H}^{T}}{{\bf ZH}{\bf H}^{T}},
\end{equation}
where $\hdp$ is the Hadamard product (i.e., element-wise product) of matrices. The multiplicative update rules have been widely adopted in NMF as they are easy to implement
and also exhibit satisfactory trade-off between the accuracy and the speed of convergence.
The NMF algorithm based on the multiplicative update rules \eqref{eq:MU_H} and \eqref{eq:MU_Z}, which is denoted as MU-NMF,   can be summarized as follows:
\begin{enumerate}
\item Initialize $\bf Z$ and $\bf H$ such that both are element-wisely positive.

\item Update $\bf H$ using  \eqref{eq:MU_H} with fixed $\bf Z$.

\item Update $\bf Z$ using \eqref{eq:MU_Z} with fixed $\bf H$.

\end{enumerate}
Repeat the above steps 2) and 3) until convergence. More details about MU-NMF can be found in \cite{lee2001algorithms}.

\subsection{The \emph{ns}NMF Method}
Arguing that standard NMF often fails to give really parts-based representation if without additional regularizations, Pascual-Marqui et al.  proposed the following non-smooth NMF (\emph{ns}NMF)  model \cite{pascual2006nonsmooth}
\begin{equation}\label{eq_6}
\mat{X}=\mat{ZSH}, \; \mat{Z}\ge \matO, \mat{H}\ge \matO,
\end{equation}
where matrices $\bf X$, $\bf Z$ and $\bf H$ are the same as those in NMF. The positive symmetric matrix ${\bf S}\in{\bf \Re}_{+}^{r\times r}$ is a ``smoothing'' matrix defined as:
\begin{equation}\label{eq_7}
{\bf S}=\left(1-\theta\right){\bf I}+\frac{\theta}{r}{{\bf 11}^{T}}
\end{equation}
where $\bf I$ is the identity matrix, $\bf 1$ is a vector of ones, and the parameter $\theta$ satisfies $0\le\theta\le 1$.

The degree of smoothness of $\bf S$ is controlled by the parameter $\theta$. As $\theta\rightarrow 1$, the matrix $\bf S$ becomes more smooth as it tends to be a constant matrix with all elements almost equal. Note that a more smooth matrix $\bf S$ will enforce stronger sparseness in both $\bf Z$ and  $\bf H$ in order to maintain faithful fit of the model to the data. In other words, the parameter $\theta$  controls the sparseness of both $\bf Z$ and $\bf H$.

%

\section{ Deep nsNMF for Hierarchy Feature Extraction}

\subsection{The Model}\label{abstract feature description}
\begin{figure}
\centering
\includegraphics[width=0.2\textwidth]
{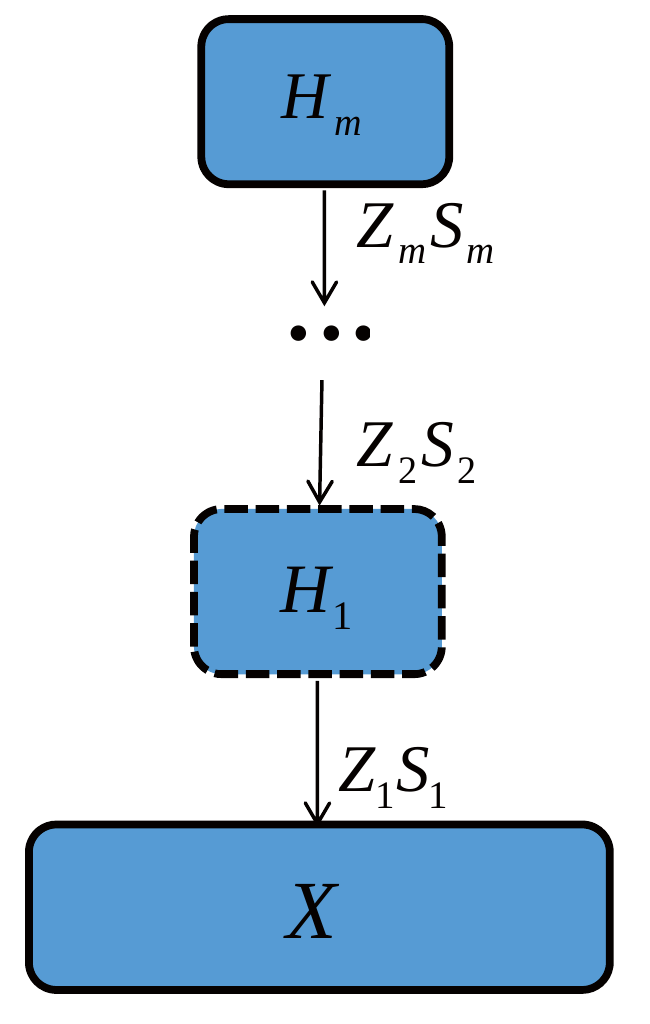}
\caption{Deep \emph{ns}NMF learns a hierarchy of different level features that aid in uncovering the final high level feature representation of the data.}
\label{archi}
\end{figure}

The goal of \emph{ns}NMF is to find localized and less overlapped features of  data by using a relatively large value of $\theta$. As a result, \emph{ns}NMF always tends to learn low level features such as contours, corners, angles, and surface boundaries. This is just the motivation behind the \emph{ns}NMF method. On the contrary, when $\theta=0$, \emph{ns}NMF boils down to ordinary NMF and may give non-sparse, largely overlapped features.
 In many practical applications, the data we wish to analyze is rather complex and is often produced by a mix model with many variables and factors. In order to better understand such complex data, it becomes compelling to disentangle the variables and  extract low and high level features simultaneously. To fulfill this purpose,  we propose a modified \emph{ns}NMF with deep architecture (called dnsNMF). In dnsNMF, the data is represented using a multilayer structure:
\begin{equation}
\label{eq_dnsNMFmodel}
{\bf X}={\bf Z}_{1}{\bf S}_{1}{\bf Z}_{2}{\bf S}_{2}\cdots{\bf Z}_{m}{\bf S}_{m}{\bf H}_{m},
\end{equation}
which generates a total of $m$ layers of features given by
\begin{eqnarray}\label{eq:dnsNMF_fea}
{\bf W}_{1}&=& {\bf Z}_{1},\nonumber
\\{\bf W}_{2}&=& {\bf Z}_{1}{\bf S}_{1}{\bf Z}_{2},\nonumber
\\&\vdots&\nonumber
\\{\bf W}_{m}&=& {\bf Z}_{1}{\bf S}_{1}{\bf Z}_{2}{\bf S}_{2}\cdots{\bf Z}_{m},
\end{eqnarray}
each of which is expected to develop a different level of abstraction.
The $m$ layers of features reflect the hierarchy and structural information contained in $\bf X$.
In the meanwhile, ${\bf W}_{1}$ is the lowest level features, denoting largely fragmental and very concrete features of objects, and ${\bf W}_{m}$ is in the highest level and denotes more meaningful and  abstract features. Moreover,  from \eqref{eq:dnsNMF_fea} we have
\begin{equation}
\label{eq_wisz}
\mats[i]{W}=\mats[i-1]{W}\mats[i]{S}\mats[i]{Z}, \; i=2,3,\ldots,m,
\end{equation}
implying that higher-level features are additive combinations of lower-level features. Particularly,  \eqref{eq_dnsNMFmodel} can be viewed as a cascade of \emph{ns}NMF (See \figurename \ref{archi}).  Taking the power of both \emph{ns}NMF and deep learning into account, dnsNMF is expected to be able to: (1) Automatically learn hierarchical features, where the higher level features are the combination of lower level features; (2) Achieve a high degree of sparseness in both the feature factors and feature representation, which largely retains the parts-based learning ability of \emph{ns}NMF. To understand the intuition of  dnsNMF, features extracted by a 3-layer dnsNMF by using the CBCL database are shown in \figurename \ref{fig_dnsnmf}. It can be seen from \figurename \ref{fig_dnsnmf}(b) that, features in the first layer mostly are pixels and outlines. In the second layer, connected edges and  contours of facial parts (nose, lips, eyebrow) appear, as shown in \figurename \ref{fig_dnsnmf}(c). In the third layer, some entire facial parts like nose, lips, and eyebrow are visibly presented, see  \figurename \ref{fig_dnsnmf}(d).

\subsection{Nesterov-type Optimization for dnsNMF}
The Deep \emph{ns}NMF algorithm consists of two key stages: pre-training and fine-tuning. In the pre-training stage, the original data matrix ${\bf X}$ is decomposed using the \emph{ns}NMF model, i.e., ${\bf X} \approx {\bf Z}_{1}{\bf S}_{1}{\bf H}_{1}$, where ${\bf Z}_{1}\in{\bf \Re}_{+}^{p\times r_{1}}$, ${\bf S}_{1}\in{\bf \Re}_{+}^{r_{1}\times r_{1}}$ and ${\bf H}_{1}\in{\bf \Re}_{+}^{r_{1}\times n}$. Then,  the representation matrix is again factorized using  \emph{ns}NMF such that  ${\bf H}_{1}= {\bf Z}_{2}{\bf S}_{2}{\bf H}_{2}$, where ${\bf Z}_{2}\in{\bf \Re}_{+}^{r_{1}\times r_{2}}$, ${\bf S}_{2}\in{\bf \Re}_{+}^{r_{2}\times r_{2}}$ and ${\bf H}_{2}\in{\bf \Re}_{+}^{r_{2}\times n}$. This procedure will continue till a desired depth has been reached.

In the fine-tuning stage, the factor matrices in all layers will be fine-tuned using a global-like minimization. In detail, the following optimization problem will be solved:
\begin{eqnarray}\label{eq:dnsNMFobj}
&&E_{Deep}{\left({\bf Z}_{1},{\bf Z}_{2},\cdots,{\bf Z}_{m},{\bf H}_{m}\right)}
\nonumber
\nonumber\\
&&=\frac{1}{2}\|{\bf X}-{\bf Z}_{1}{\bf S}_{1}{\bf Z}_{2}{\bf S}_{2}\cdots{\bf Z}_{m}{\bf S}_{m}{\bf H}_{m}\|_{F}^{2}.
\end{eqnarray}
Apparently, the problem is highly non-convex and it is unrealistic to expect to find global solutions. Taking into account that the cost function is convex with respect to each single factor matrix,  a common practice to solve this problem is applying the block coordinate descend method, i.e., minimizing the cost function with respect to only one factor matrix each time, while remaining the others fixed. Note that every factor matrix must be updated at least once after a certain number of iterations, which is often referred to as a sweep. Below we discuss how to update each factor matrix.



$\bullet\;$\emph{Update ${\bf Z}_{1}$ and ${\bf H}_{m}$.} This is the simplest case as these two cases actually have been extensively studied in nonnegative matrix/tensor factorization. When all factor matrices but \mats[m]{H} are fixed, problem  \eqref{eq:dnsNMFobj} is equivalent to the following optimization problem:
\begin{eqnarray}
\label{eq:dnsNMFobj_hm}
\min_{\mats[m]{H}\ge\matO} \; E_{Deep}{\left({\bf H}_{m}\right)}=\frac{1}{2}\|{\bf X}-{\bf A}{\bf H}_{m}\|_{F}^{2}.
\end{eqnarray}
where
\begin{equation}
\label{eq:A4Hm}
{\bf A} = {\bf Z}_{1}{\bf S}_{1}{\bf Z}_{2}{\bf S}_{2}\cdots{\bf Z}_{m}{\bf S}_{m}.
\end{equation}
Problem \eqref{eq:dnsNMFobj_hm} is a standard nonnegative least squares problem, and any state-of-the-art methods in \cite{SPM_NMFNTD, NECO2016NTF, guan2012nenmf} can be applied. As all these methods are essentially the first-order methods, we are particularly interested in the accelerated proximal gradient (APG) method, which is often viewed as the optimal first-order method. While the details of APG based NMF/NTF  can be found in \cite{NECO2016NTF, guan2012nenmf}, we briefly introduce the APG method here to make paper self-contained.

\begin{lemma}[ \cite{NECO2016NTF, guan2012nenmf}]
\label{th:apgHm}
The gradient  of $E_{deep}(\mats[m]{H})$, i.e.,
\begin{equation}
\label{eq:gradHm}
\triangledown_{ \mats[m]{H}}E_{deep}{\left(\mats[m]{H}\right)}=\trans{\mat{A}}\mat{A}\mats[m]{H}-\trans{\mat{A}}\mat{X},
\end{equation}
 is Lipschitz continuous and $L=\frob[2]{\trans{\mat{A}}\mat{A}}$ is a Lipschitz constant.
\end{lemma}
We briefly explain the core idea of the APG method below. For notational simplicity we temporarily define
\begin{equation}
g(\mats[m]{H})\defeq E_{deep}{\left(\mats[m]{H}\right)}.
\end{equation}
Then, the Lipschitz constant $L$ ensures that, $\forall \mat{H}, \tilde{\mat{H}}$,
\begin{equation}
\frob{g'{\left(\mat{H}\right)}-g'{\left(\tilde{\mat{H}}\right)}}\le L\frob{\mat{H}-\tilde{\mat{H}}},
\end{equation}
thereby leading to \cite[Appendix A.24]{NP_BCD}
\begin{equation}
\begin{split}
g(\mat{H})\le & \phi(\mat{H}) \\
\defeq& g(\tilde{\mat{H}})+\innerprod{g'(\tilde{\mat{H}})}{\mat{H}-{\tilde{\mat{H}}}}
+\frac{L}{2}\frob{\tilde{\mat{H}}-\mat{H}}^{2}.
\end{split}
\end{equation}
The APG method minimizes the upper bound $\phi(\mat{H})$ instead of $g(\mat{H})$ directly. Note that $\phi(\mat{H})$ is convex w.r.t. \mat{H} and the global optimum is $\proj_{+}\left(\tilde{\mat{H}}-\frac{1}{L}g'(\tilde{\mat{H}})\right)$, where the operator $\proj_{+}(\cdot)$ sets all negative entries to be zeros. Hence, we optimize $g(\mats[m]{H})$ by updating \mats[m]{H} iteratively using
\begin{equation}
\label{eq:apgiter}
\mats[m]{H}\from \proj_{+}\left(\tilde{\mat{H}}-\frac{1}{L}g'(\tilde{\mat{H}})\right).
\end{equation}
Let $\mats[m]{H}^{k}$ be the value of \mats[m]{H} after $k$ iterations. It can be verified that
\begin{enumerate}
\item
if $\tilde{\mat{H}}=\mats[m]{H}^{k}$, \eqref{eq:apgiter} leads to a projected gradient method with the convergence rate no worse than \bigO{1/k}.
\item
if $\tilde{\mat{H}}=\mats[m]{H}^{k}+\frac{\alpha_{k-1}-1}{\alpha_{k}}\left(\mats[m]{H}^{k}-\mats[m]{H}^{k-1}\right)$, where the sequence \set{\alpha_{k}} satisfies $\alpha_{k}^{2}-\alpha_{k}\le\alpha_{k-1}^{2}$, \eqref{eq:apgiter} leads to a very fast convergence rate of \bigO{1/k^{2}}. See \cite{guan2012nenmf, FISTA} for detailed discussion.
\end{enumerate}
In summary, problem \eqref{eq:dnsNMFobj_hm} can be efficiently solved by using the APG method, see  Algorithm \ref{alg:APG} for the pseudo-code.

Similarly, by fixing all factors but \mats[1]{Z}, problem  \eqref{eq:dnsNMFobj} is equivalent to
\begin{eqnarray}
\label{eq:dnsNMFobj_z1}
{\bf Z}_{1} = \text{arg min}_{{\bf Z}_{1}} E_{Deep}(\mats[1]{Z})=\frac{1}{2}\|{\bf X}^{T}-{\bf B}^{T}{\bf Z}_{1}^{T}\|_{F}^{2},
\end{eqnarray}
where ${\bf B} = {\bf S}_{1}{\bf Z}_{2}{\bf S}_{2}\cdots{\bf Z}_{m}{\bf S}_{m}{\bf H}_{m}$, and again it can be solved by applying Algorithm \ref{alg:APG} similarly.

\begin{algorithm}[t]
\caption{The APG Algorithm for  $\min_{\mats[m]{H}}E_{deep}(\mats[m]{H})$.}
\label{alg:APG}
\begin{algorithmic}[1]
\REQUIRE
           {$\mat{X}$, $\mat{A}$ given in \eqref{eq:A4Hm}}
\STATE
      {Initialize $\mats[m]{H}^k$, ${\bf Y}^{k}={\bf H}_{m}^{k}$, $\alpha_{0}=1$, $k=0$. $L=\frob[2]{\trans{\mat{A}}\mat{A}}$.}
\WHILE {not converage}
\STATE {${\bf H}_{m}^{k+1}\leftarrow \proj_{+}\left({\bf Y}^{k}-\frac{1}{L}\triangledown_{ \mats[m]{H}}E_{deep}{|}_{\mats[m]{H}=\mat{Y}^k}\right)$.}
\STATE {$\alpha_{k+1}\leftarrow\frac{1+\sqrt{4\alpha_{k}^{2}+1}}{2}$.}
\STATE {${\bf Y}^{k+1}\leftarrow \mats[m]{H}^{k+1}+\frac{\alpha_{k}-1}{\alpha_{k+1}}(\mats[m]{H}^{k+1}-\mats[m]{H}^{k})$.}
\STATE {$k\leftarrow k+1$.}
\ENDWHILE
\RETURN {$\mats[m]{H}={\bf H}_{m}^{k}$.}
\end{algorithmic}
\end{algorithm}

$\bullet\;$\emph{Update $\mats[i]{Z}$ for $ i=2, 3, \cdots$. } By fixing all factors but \mats[i]{Z}, $i>1$, we obtain
\begin{eqnarray}
\label{eq:dnsNMFobj_zi}
{\bf Z}_{i} = \text{arg min}_{{\bf Z}_{i}} E_{Deep}{\left(\mats[i]{Z}\right)}=\frac{1}{2}\|{\bf X}-{\bf A}_{i}{\bf Z}_{i}{\bf B}_{i}\|_{F}^{2},
\end{eqnarray}
where
\begin{equation}
\begin{split}
\mats[i]{A} &=\mats[1]{Z}\mats[1]{S}\cdots\mats[i-1]{Z}\mats[i-1]{S},  \nonumber\\
\mats[i]{B} &=\mats[i]{S}\mats[i+1]{Z}\cdots\mats[m]{S}\mats[m]{H}, \; i>1.\nonumber
\end{split}
\end{equation}
In order to optimize \mats[i]{Z} using the APG method, we need to show that the gradient $\triangledown_{ \mats[i]{Z}}E_{deep}(\mats[i]{Z})$ is also Lipschitz continuous.
Note that the cost function $E_{Deep}{\left(\mats[i]{Z}\right)}$ can be rewritten in a vector form as
\begin{equation}
\label{eq:dnsNMFobj_zivec}
E_{Deep}{\left(\mats[i]{Z}\right)}=\|{\vecit{\bf X}}-(\trans{\mats[i]{B}}\kkp\mats[i]{A})\vecit{\mats[i]{Z}}\|_{F}^{2},
\end{equation}
where $\vecit{{\bf Z}_{i}}$ is the vectorization operator of the matrix ${\bf Z}_{i}$, and $\kkp$ is the Kronecker product of matrices. As such, \eqref{eq:dnsNMFobj_zivec} possesses exactly the same form as \eqref{eq:dnsNMFobj_hm} and \eqref{eq:dnsNMFobj_z1}, and can be minimized similarly.
By following the analysis in \cite{NECO2016NTF,lraNTD}, we have the following proposition:


\begin{proposition}
The gradient of  $E_{Deep}{\left({\mats[i]{Z}}\right)}$, i.e.,
\begin{equation}
\label{eq:gradZi}
\triangledown_{ \mats[i]{Z}}E_{deep}(\mats[i]{Z})=\trans{\mats[i]{A}}\mats[i]{A}\mats[i]{Z}\mats[i]{B}\trans{\mats[i]{B}}-\trans{\mats[i]{A}}\mat{X}\trans{\mats[i]{B}},
\end{equation}
 is Lipschitz continuous and $L_{i}=\|{\bf A}_{i}^{T}{\bf A}_{i}\|_{2}\|{\bf B}_{i}{\bf B}_{i}^{T}\|_{2}$ is a Lipschitz constant.
\end{proposition}
The proof is rather straightforward and can be found in Appendix A. Due to \eqref{eq:gradZi} and Proposition 1, the matrices \mats[i]{Z} can be updated using the APG method efficiently.

We summarize the dnsNMF algorithm in Algorithm \ref{alg:dnsNMF}.

\renewcommand{\algorithmicensure}{\textbf{Output:}}
\begin{algorithm}[t]
\caption{The dnsNMF algorithm.}
\label{alg:dnsNMF}
\begin{algorithmic}[1]
\REQUIRE
       {${\bf X}\in{\bf\Re}^{p\times n}$ and the layer sizes $[r_{1}, r_{2},\cdots, r_{m}]$
\ENSURE
       {Feature factors ${\bf Z}_{i}$ and feature presentation ${\bf H}_{m}$ such that ${\bf X}\approx {\bf Z}_{1}{\bf S}_{1}{\bf Z}_{2}{\bf S} _{2}\cdots{\bf Z}_{m}{\bf S}_{m}{\bf H}_{m}$, where $\mats[i]{Z}\in\Real^{r_{i-1}\times r_i}_+$, $1\le i\le m$, , $r_0=p$, and $\mats[m]{H}\in\Real^{r_{m}\times n}_+$.}}

 \STATE{ Initialize for $t=1$:
  \begin{equation}
 \begin{split}
( {\bf Z}_{1}^{t},{\bf S}_{1}^{t},{\bf H}_{1}^{t})&\leftarrow\emph{ns}\textrm{NMF}\left({\bf X}\right);\nonumber
 \\({\bf Z}_{i}^{t},{\bf S}_{i}^{t},{\bf H}_{i}^{t})&\leftarrow\emph{ns}\textrm{NMF}\left({\bf H}_{i-1}^{t}\right), \text{for}\; i=2,3,\ldots,m.\nonumber
 \end{split}
 \end{equation}
 }

 \REPEAT
 \STATE ${\bf B} = {\bf S}_{1}{\bf Z}_{2}^{t}{\bf S}_{2}\cdots{\bf Z}_{m}^{t}{\bf S}_{m}{\bf H}_{m}^{t}$.
 \STATE ${\bf Z}_{1}^{t+1}\from\arg\min_{\mats[1]{Z}} \frac{1}{2}\frob{\trans{\mat{X}}-\trans{\mat{B}}\trans{\mats[1]{Z}}}^2$ by using, say, the APG method.

 \FOR{$i=2$ \TO $m$}
 \STATE{$\mats[i]{A} =\mats[1]{Z}^{t+1}\mats[1]{S}\cdots\mats[i-1]{Z}^{t+1}\mats[i-1]{S}$.}
\STATE{$\mats[i]{B} =\mats[i]{S}\mats[i+1]{Z}^{t}\cdots\mats[m]{S}\mats[m]{H}^{t}$.}
\STATE{$\mats[i]{Z}^{t+1}\from\arg\min_{\mats[i]{Z}} \frac{1}{2}\frob{\mat{X}-\mats[i]{A}\mats[i]{Z}\mats[i]{B}}^2$.}
 \ENDFOR

 \STATE{${\bf A} = {\bf Z}_{1}^{t+1}{\bf S}_{1}{\bf Z}_{2}^{t+1}{\bf S}_{2}\cdots{\bf Z}_{m}^{t+1}{\bf S}_{m}$.}
 \STATE{${\bf H}_{m}^{t+1}\from\arg\min_{\mats[m]{H}} \frac{1}{2}\frob{\mat{X}-\mat{A}\mats[m]{H}}^2$.}
 \STATE{$t\from t+1$.}
 \UNTIL{A stopping criterion is met}
\end{algorithmic}
\end{algorithm}

\section{Relationship Between dnsNMF With Autoencoder}
\begin{figure}
\centerline{
\subfloat[Autoencoder]{\includegraphics[width=0.35\textwidth]
{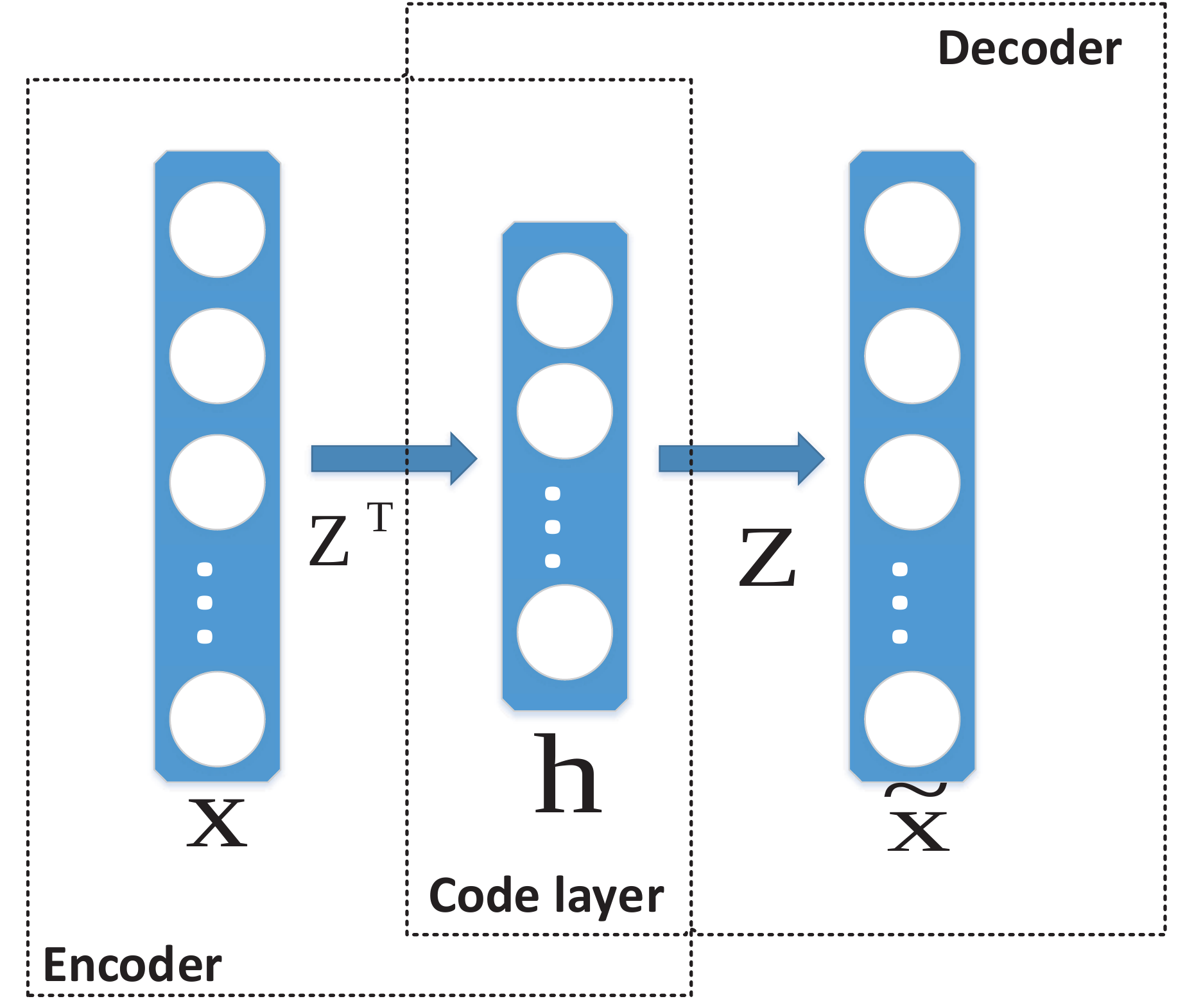}}
}
\centerline{
\subfloat[Deep Autoencoder]{\includegraphics[width=0.35\textwidth]
{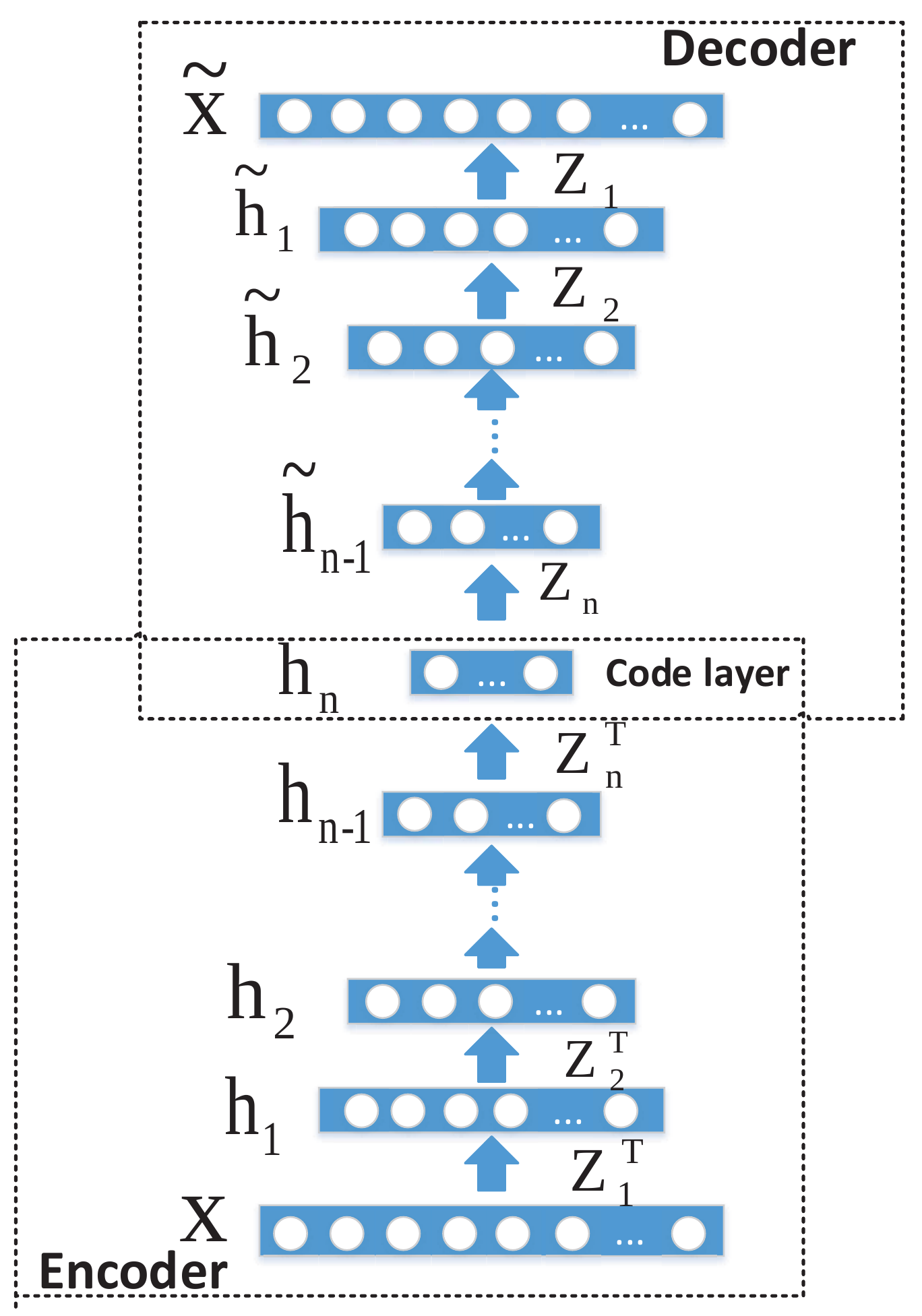}}
}
\caption{Illustration of autoencoder and deep autoencoder}
\label{fig:ae}
\end{figure}

It is well known that autoencoders  \cite{munro1988principal} may be thought of as a special case of feedforward networks, which are able to extract representation of data with significantly reduced dimensionality. The structure of an autoencoder is shown in \figurename \ref{fig:ae}(a). Typically, an autoencoder consists of two parts: an encoder function ${\bf h}=f(\bf x)$ that transforms the high dimensional data into a low-dimensional code, and a decoder that produces the reconstruction ${\bf\tilde{x}}=g(\bf h)$ that approximates the original data from the output of the encoder. The two parts are generally pre-trained together by minimizing the discrepancy between the original data and its reconstruction \cite{hinton2006reducing}.

To create a deep autoencoder, multiple layers of encoder and decoder will be stacked in order, as illustrated in \figurename \ref{fig:ae}(b). In the encoding stage, the output of each layer will be the input of its neighbored upper layer, and the dimensionality of outputs will be smaller and smaller, implying a more abstract and more compact representation of the original data. In the decoding stage, the encoding layers are ``unrolled'' to reconstruct the original data.


It can be seen that dnsNMF has the same structure as autoencoder. In the encoding stage, data are compressed with both nonnegativity and sparsity constrained feature representations and weights, which preserves the learning parts ability of NMF. Due to the nonnegativity nature of outputs of \emph{ns}NMF, we may assume that the rectified linear units (ReLU) are applied in the sense that $g(\mats[l]{Z}\mats[l]{S}\mats[l]{h})=\mats[l]{Z}\mats[l]{S}\mats[l]{h}$, where  $g(x)=\max(0,x)$ is the rectified linear activation function. As such, dnsNMF can be represented as
\begin{eqnarray}\label{eq_decoder}
\qquad{\bf\tilde{x}}&=& g({\bf Z}_{1}\mats[1]{S}{\bf\tilde{h}}_{1}),\nonumber
\\{\bf\tilde{h}}_{1}&=& g({\bf Z}_{2}\mats[2]{S}{\bf\tilde{h}}_{2}),\nonumber
\\&\vdots&\nonumber
\\{\bf\tilde{h}}_{m-1}&=& g({\bf Z}_{m}\mats[m]{S}{\bf h}_{m}).
\end{eqnarray}
or equivalently in a compact form,
\begin{eqnarray}\label{eq_decoder_2}
{\bf x}\approx {\bf\tilde{x}}=g({\bf Z}_{1}\mats[1]{S}g({\bf Z}_{2}\cdots g({\bf Z}_{m}\mats[m]{S}{\bf h}_{m}))),
\end{eqnarray}
implying that dnsNMF has the same structure as autoencoder.

Certainly, it also should be note that there are some differences between dnsNMF and deep autoencoder. dnsNMF is a unsupervised learning method which does not need data labels, but deep autoencoder is a semi-supervised learning method which uses unsupervised learning method to pre-train network and utilizes the supervised learning method to fine-tune it. dnsNMF, as an unsupervised learning method, can be applied to explore the unknown hierarchical features hidden in data while deep autoencoder are often be used to extract the features which are related to the data labels. From this point, they play different roles in data analysis and it is not significant to tell which is the better method.


\section{Simulations and Experiments}
In this section, the proposed dnsNMF method was compared with related state-of-the-art methods by performing clustering analysis on three human face data sets, i.e. the ORL data set, the Jaffe data set, and the Yale data set. We also demonstrated abilities of the dnsNMF to extract sparse hierarchical features of data by using the CBCL data set.
\subsection{Data Sets}
Four data sets were used in our  experiments:
\begin{itemize}
 \item AT\&T ORL\footnote{Available at \url{http://www.uk.research.att.com/facedatabase.html.}}: The AT\&T ORL database consists of 10 different images for each of 40 distinct subjects, thus 400 images in total. The images were taken at different time with varying lighting, facial expressions (open/closed eyes, smiling/not smiling), and facial details (glasses/no glasses). All the images were taken against a dark homogeneous background with the subjects in an upright, frontal position.

 \item JAFFE\footnote{Available at \url{http://www.kasrl.org/jaffe_info.html}}: The database contains 213 images of 7 facial expressions (6 basic facial expressions + 1 neutral) posed by 10 Japanese female models. Each image has been rated on 6 emotion adjectives by 60 Japanese subjects. The database was designed and assembled by Michael Lyons, Miyuki Kamachi, and Jiro Gyoba. See for more details \cite{lyons1998coding}.
 \item Yale\footnote{Available at \url{http://cvc.yale.edu/projects/yalefaces/yalefaces.html.}}: The Yale database contains 165 grayscale images in GIF format of 15 individuals. There are 11 images per subject, one per different facial expression or configuration: center-light, w/glasses, happy, left-light, w/no glasses, normal, right-light, sad, sleepy, surprised, and wink.
 \item CBCL Face Database\footnote{Available at \url{http://cbcl.mit.edu/cbcl/software-datasets/FaceData2.html}}: CBCL Face Database is a database of faces and non-faces, that has been extensively used at the center for Center for Biological and Computation Learning at MIT. Here, we choose the training set face, which contains 2,429 faces. Assuming that each has only $19\times19$ pixels.
 \end{itemize}
\subsection{Compared Algorithms}
The following algorithms were compared with our dnsNMF in experiments:
\begin{itemize}
\item  	The \emph{ns}NMF method proposed in \cite{pascual2006nonsmooth}, i.e., the standard dnsNMF algorithm with a single layer.
\item 	The NMF method proposed in \cite{lee2001algorithms}.
\item 	The graph regularized NMF (GNMF) \cite{cai2008non} which encodes the geometrical information into NMF.
\item 	The Deep Semi-NMF method very recently developed in \cite{trigeorgis2017deep}.
\end{itemize}
For dnsNMF and Deep Semi-NMF, after achieving satisfactory reconstruction error, the toppest layer, i.e., \mats[m]{H}, was used as the feature representation. The clustering performance was measured by two performance indices, the Accuracy (AC) and the Normalized Mutual Information  (NMI) (details about these two metrics can be found in \cite{cai2005document}).
\begin{figure*}
\centering
\subfloat[Original images]{\includegraphics[width=0.2\textwidth,height=0.2\textwidth]
{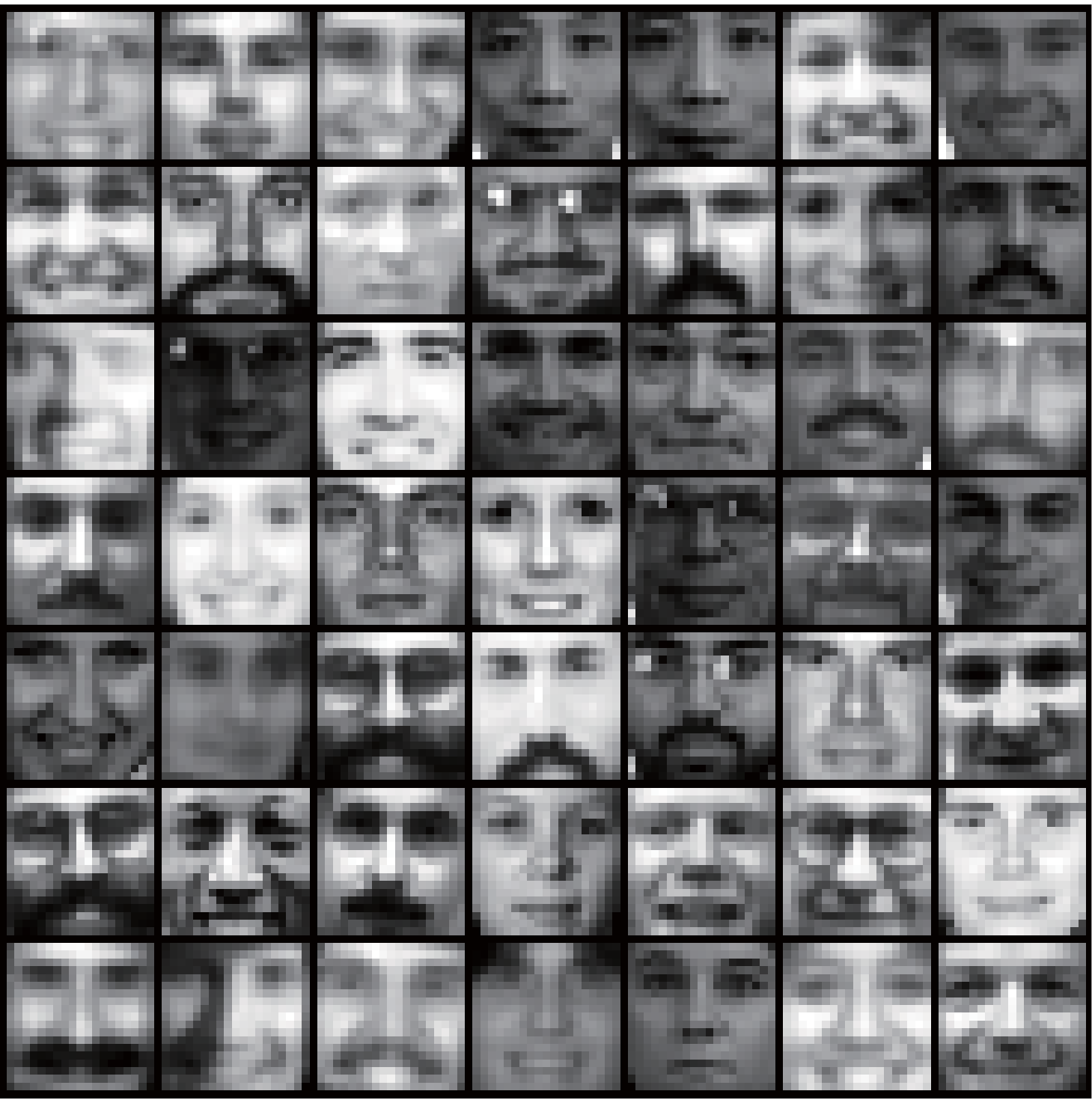}}
\subfloat[First layer features]{\includegraphics[width=0.2\textwidth,height=0.2\textwidth]
{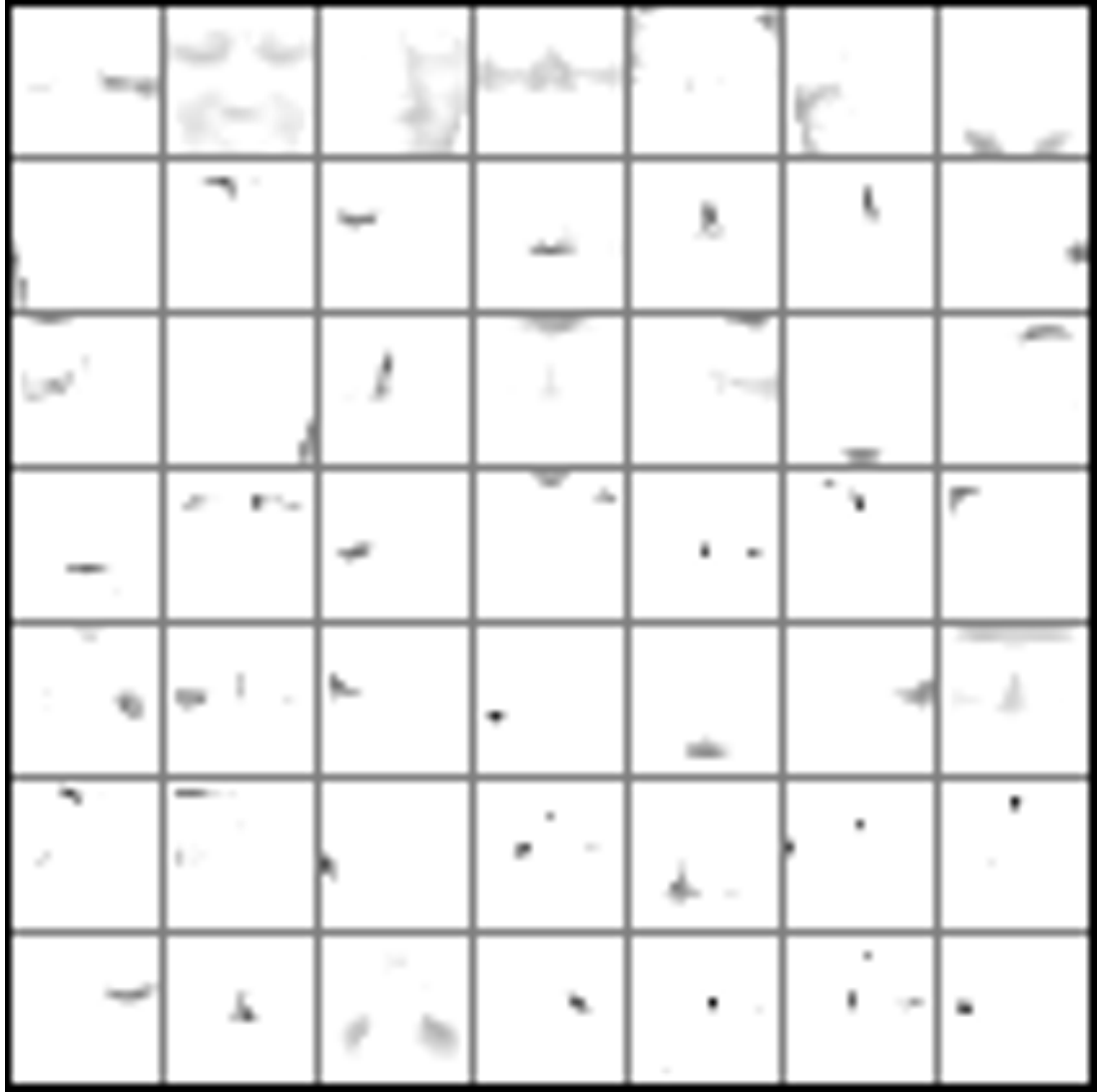}}
\subfloat[Second layer features]{\includegraphics[width=0.2\textwidth,height=0.2\textwidth]
{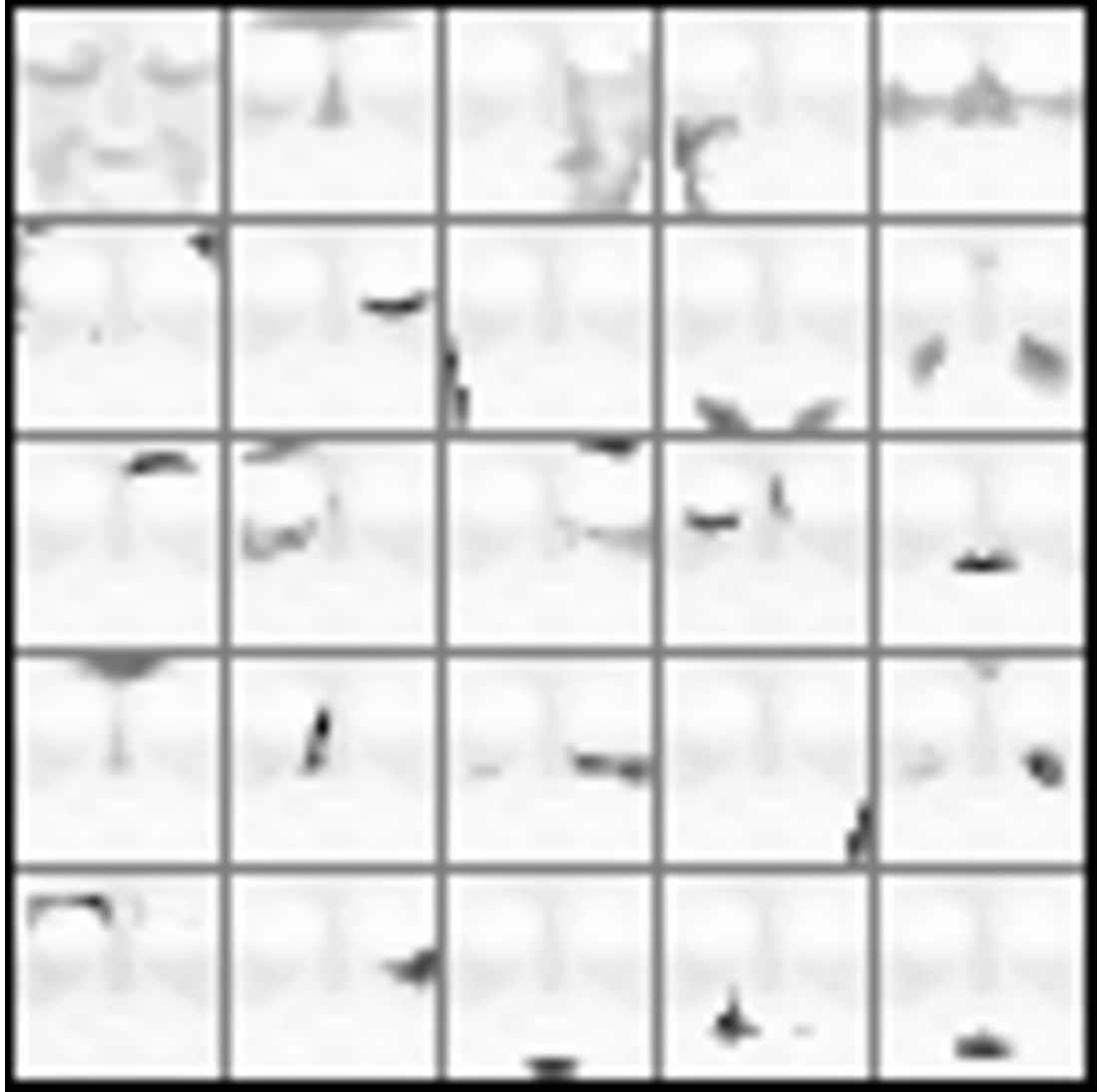}}
\subfloat[Third layer features]{\includegraphics[width=0.2\textwidth,height=0.2\textwidth]
{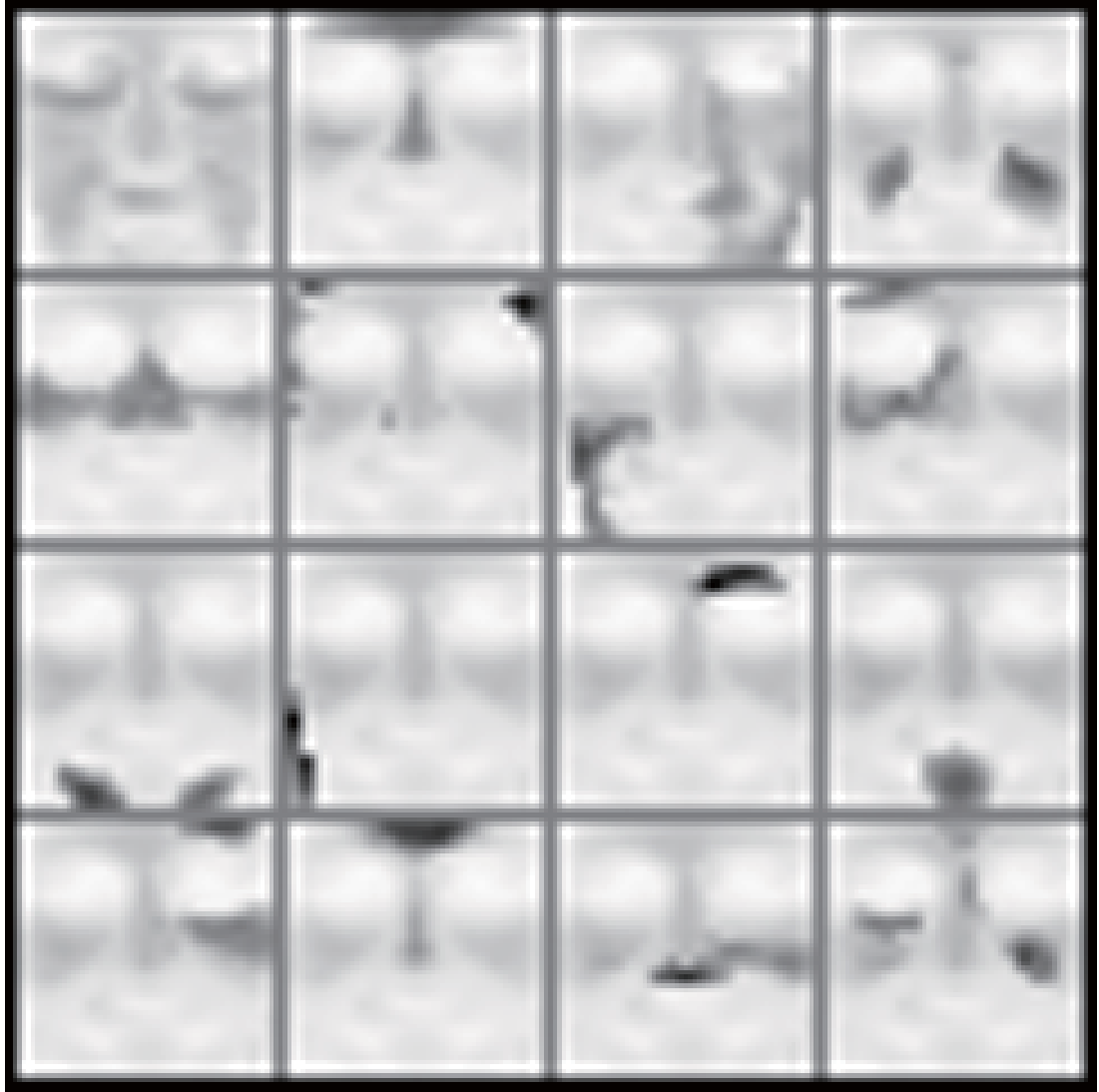}}
\caption{Features learned by dnsNMF from the CBCL face images database. (a) Original images. (b), (c), (d) show the features created by each layer, which are ${\bf Z}_{1}$,  ${\bf Z}_{1}{\bf Z}_{2}$, and ${\bf Z}_{1}{\bf Z}_{2}{\bf Z}_{3}$, respectively. }
\label{fig_dnsnmf}
\end{figure*}
\begin{figure*}
\centering
\subfloat[Original images]{\includegraphics[width=0.2\textwidth,height=0.2\textwidth]
{cbcl_originalimages0}}
\subfloat[First layer features]{\includegraphics[width=0.2\textwidth,height=0.2\textwidth]
{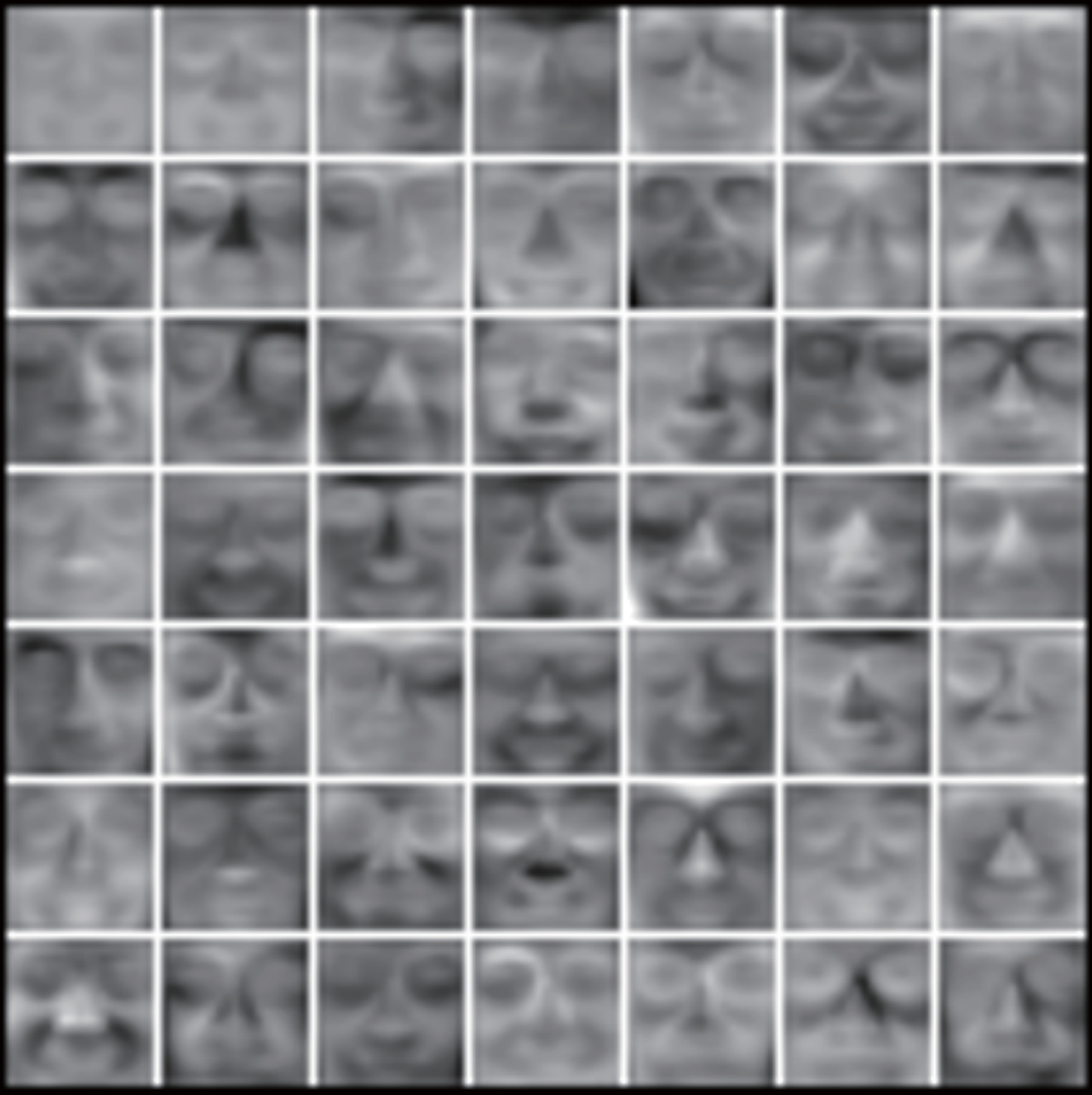}}
\subfloat[Second layer features]{\includegraphics[width=0.2\textwidth,height=0.2\textwidth]
{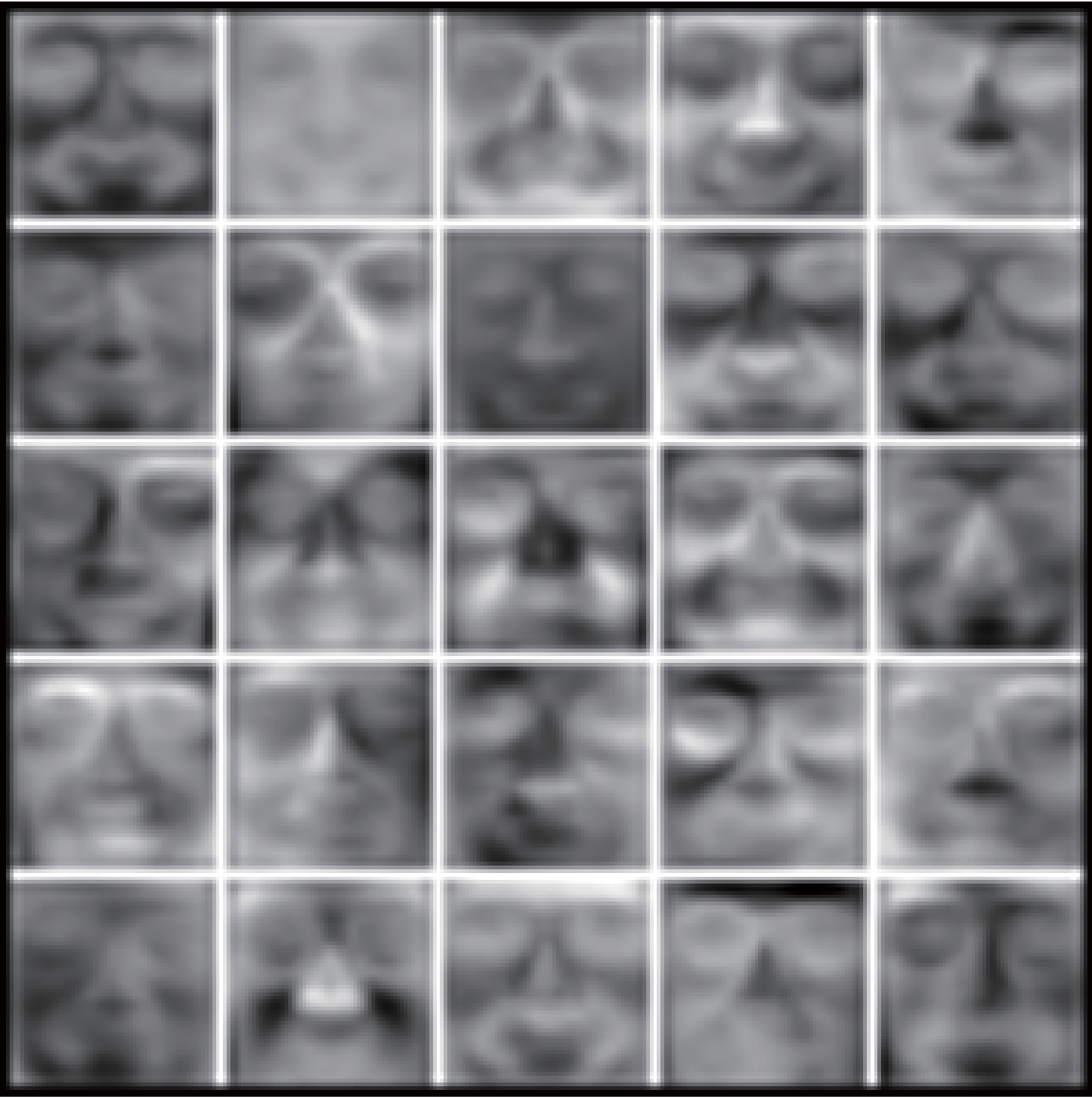}}
\subfloat[Third layer features]{\includegraphics[width=0.2\textwidth,height=0.2\textwidth]
{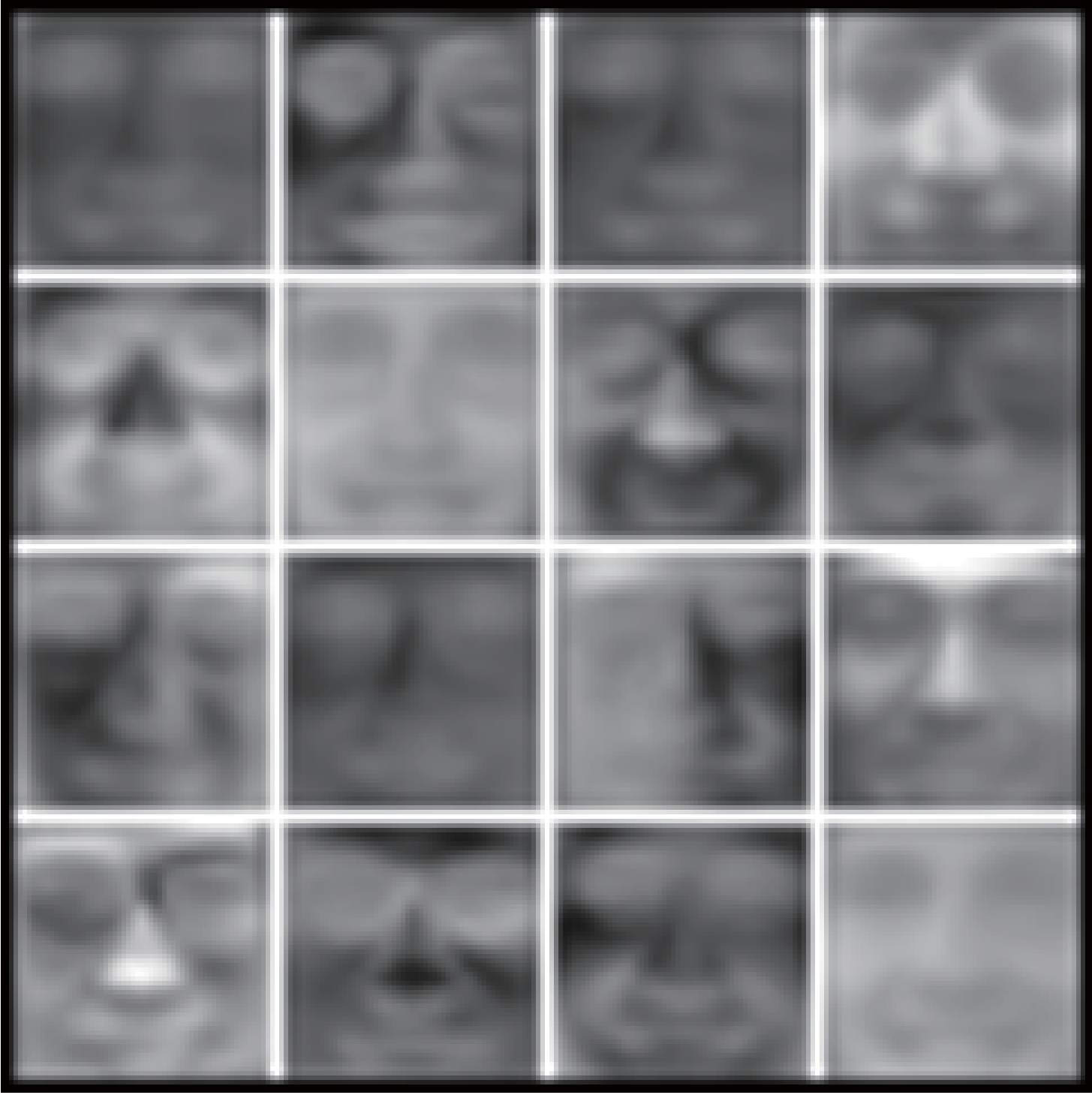}}
\caption{Features learned by Deep Semi-NMF from the CBCL face images database. (a) Original images. (b), (c), (d) show the features created by each layer, which are ${\bf Z}_{1}$,  ${\bf Z}_{1}{\bf Z}_{2}$, and ${\bf Z}_{1}{\bf Z}_{2}{\bf Z}_{3}$, respectively. }
\label{fig_dsenmf}
\end{figure*}
\begin{figure*}
\centering
\subfloat[Sparseness of presentations learned by dnsNMF]{\includegraphics[width=0.4\textwidth]
{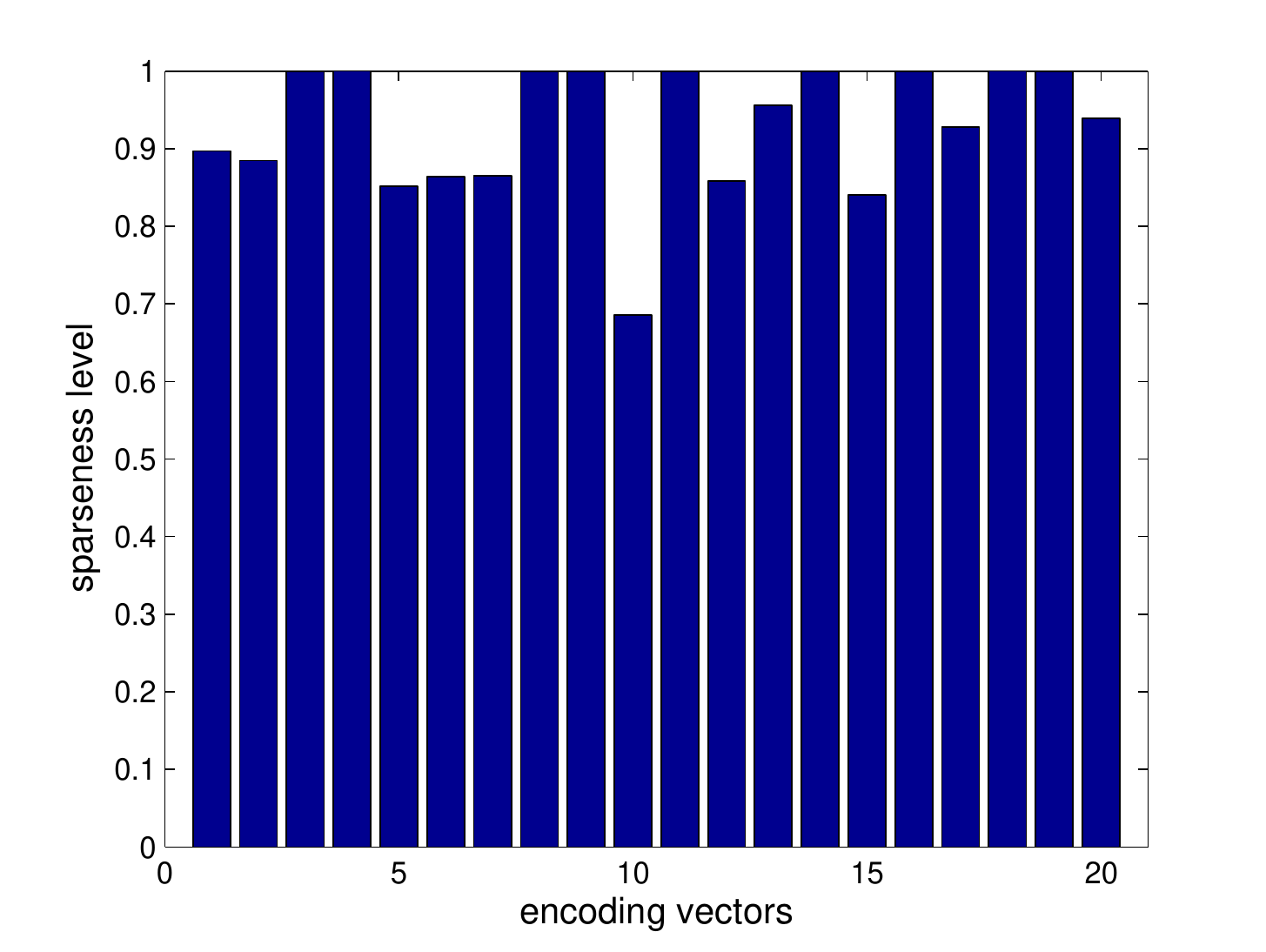}}
\subfloat[Sparseness of presentations learned by Deep Semi-NMF]{\includegraphics[width=0.4\textwidth]
{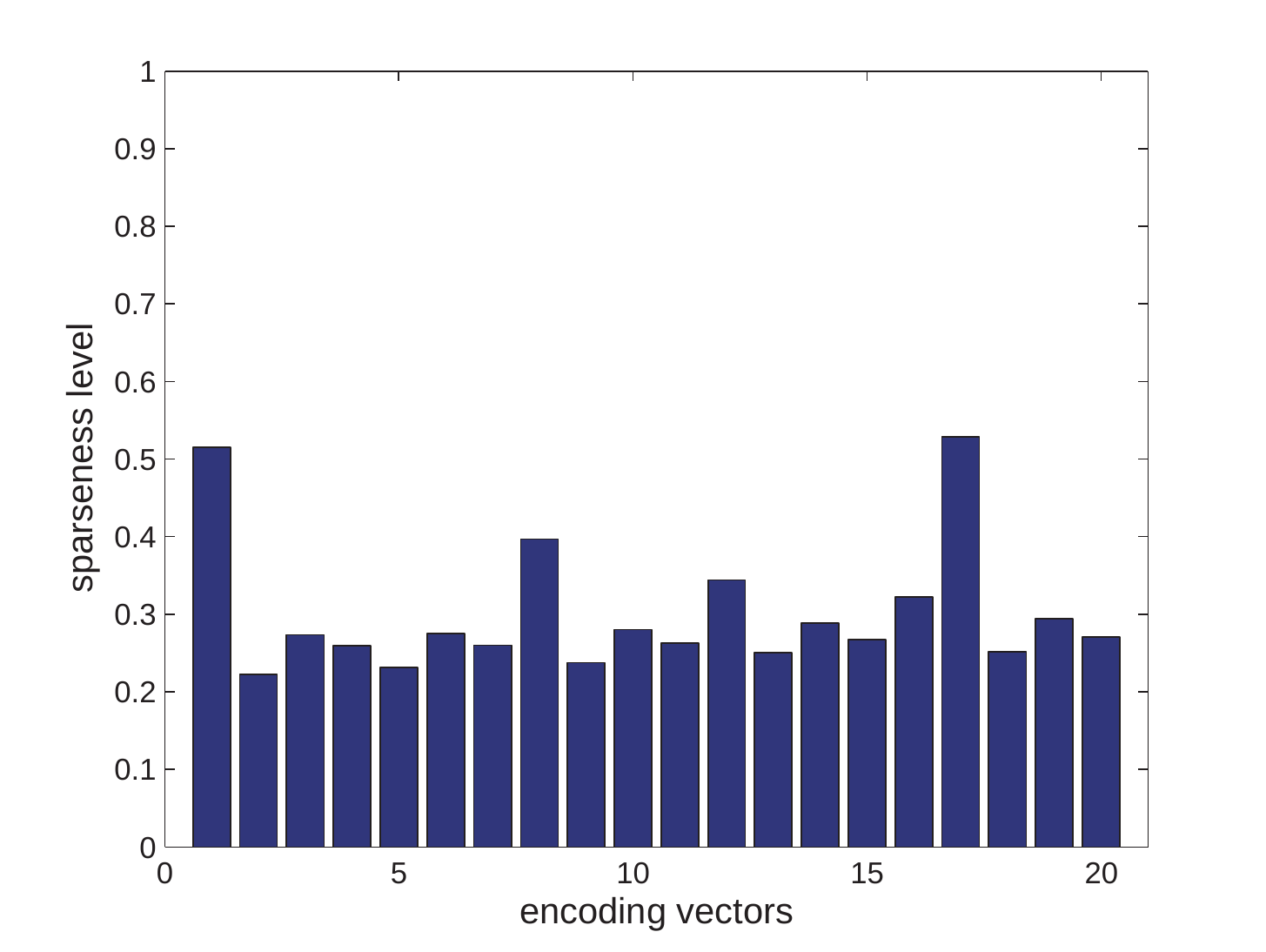}}
\caption{Illustration of the sparseness level of data representation learned from CBCL data set by dnsNMF (a) and Deep Semi-NMF (b) respectively. Here we illustrate only 20 samples chosen randomly.}
\label{fig_H}
\end{figure*}
\subsection{Visualization of Features}
In the aforementioned algorithms, only dnsNMF and Deep Semi-NMF have deep architectures. So the goal of this experiment is compare how dnsNMF and Deep Semi-NMF learn hierarchical parts-based features, respectively. In this experiment, we firstly process the CBCL face data set as did by D.D.Lee and H.S.Seung in \cite{lee1999learning} and then decompose it by using the dnsNMF and Deep Semi-NMF with three layers. The resulting basis features was visualized as $19\times 19$ gray scale images. From \figurename \ref{fig_dnsnmf}, it can be seen that dnsNMF gives pixel-level features in the first layer, and edge-based features in the second layer by combining the features generated in the previous layer, and more complex features reflecting entire facial organs in the third layer. As such, a set of parts-based hierarchical features were learnt.
\figurename \ref{fig_dsenmf} shows that counter parts learnt by the Deep Semi-NMF method. Apparently, due to the lack of full nonnegativity constraints, the learning-parts ability was largely damaged, which is one major limitation of Deep Semi-NMF when this parts-based representation plays crucial role in applications. In addition, the corresponding feature representations learned by dnsNMF and Deep Semi-NMF are shown in \figurename \ref{fig_H}(a) and (b) respectively. It can also be seen that the data representation learned by dnsNMF is much more sparse than Deep Semi-NMF.

\subsection{Experiments on Clustering Analysis of Face Images}
Similar to \cite{liu2012constrained}, for each data set the number of clusters varies from 2 to 10. For simplicity, the facial area of each image was cropped. All of the face images were gray-scaled with the size of $32\times32$ and then were normalized in scale and orientation such that the two eyes are aligned at the same position. Implementation details are as follows:
\begin{enumerate}
\item For each $k$, a total of $k$ categories from the data set were randomly chosen and shuffled as the collection of samples $\bf X$ for clustering.
\item The number of features, i.e., the number of rows of \mats[m]{H}, was selected to be equal to the number of clusters for clustering analysis for all algorithms (for those without a deep architecture, $m=1$; otherwise $m=2$).
\item K-means was applied to the extracted features \mats[m]{H} for images clustering and repeated 20 times with different initial points and the best result was recorded.
\end{enumerate}

We repeated the above procedure 10 times and recorded the average clustering performance as the final result. All the algorithms had adopted  grid search in their parameter spaces to achieve the best accuracy as possible. The dnsNMF and the Deep Semi-NMF were composed of two representation layers in all experiments. To speed up the converge of Deep Semi-NMF, as did in \cite{trigeorgis2017deep}, we used an SVD based initialization \cite{gillis2015exact}. Similarly, we used the Non-negative Double Singular Value decomposition (NNDSVD) \cite{boutsidis2008svd} to speed up dnsNMF.

\begin{figure*}
\centering
\subfloat[Accuracy vs. Number of Clusters]{\includegraphics[width=0.4\textwidth]
{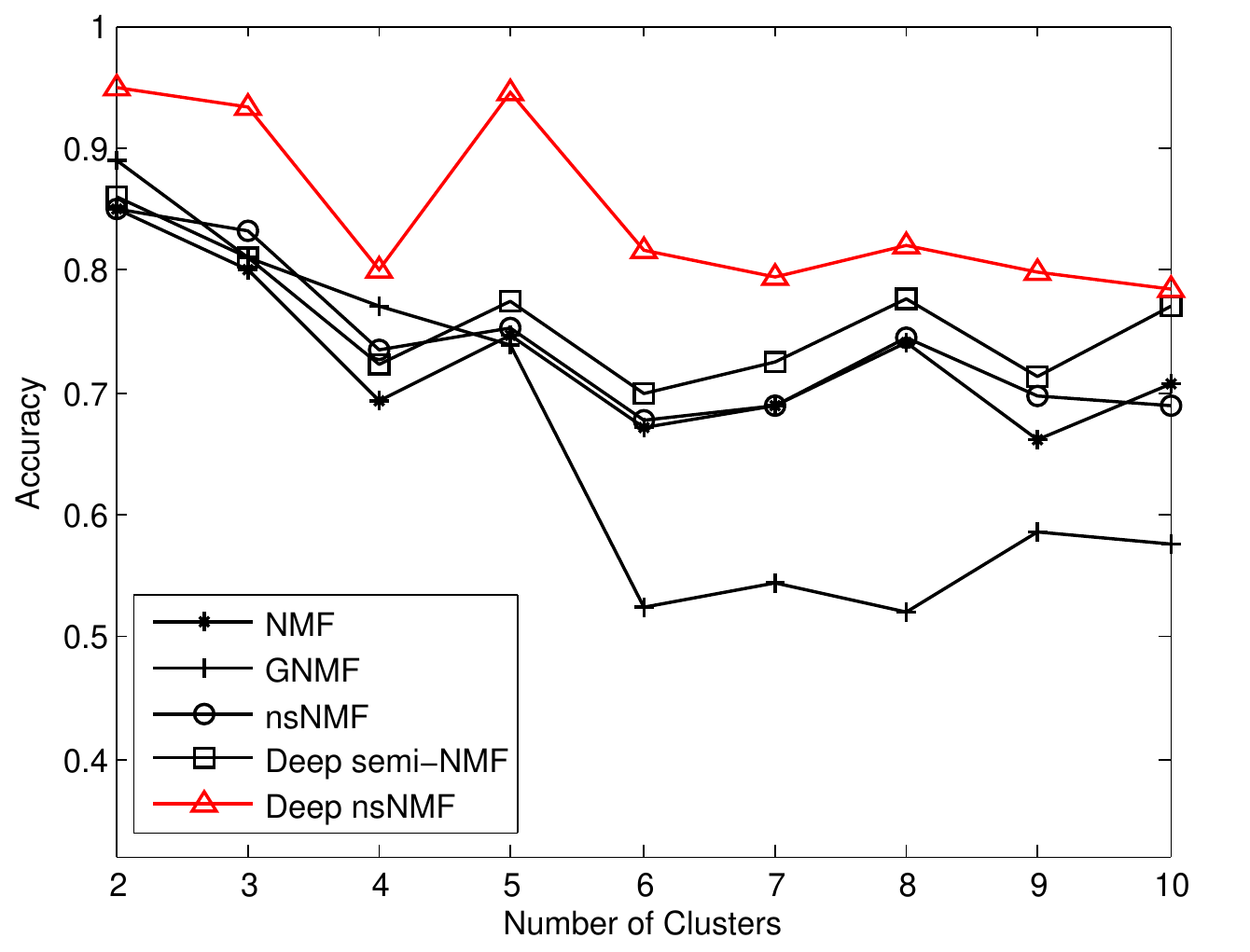}}
\subfloat[Mutual Information vs. Number of Clusters]{\includegraphics[width=0.4\textwidth]
{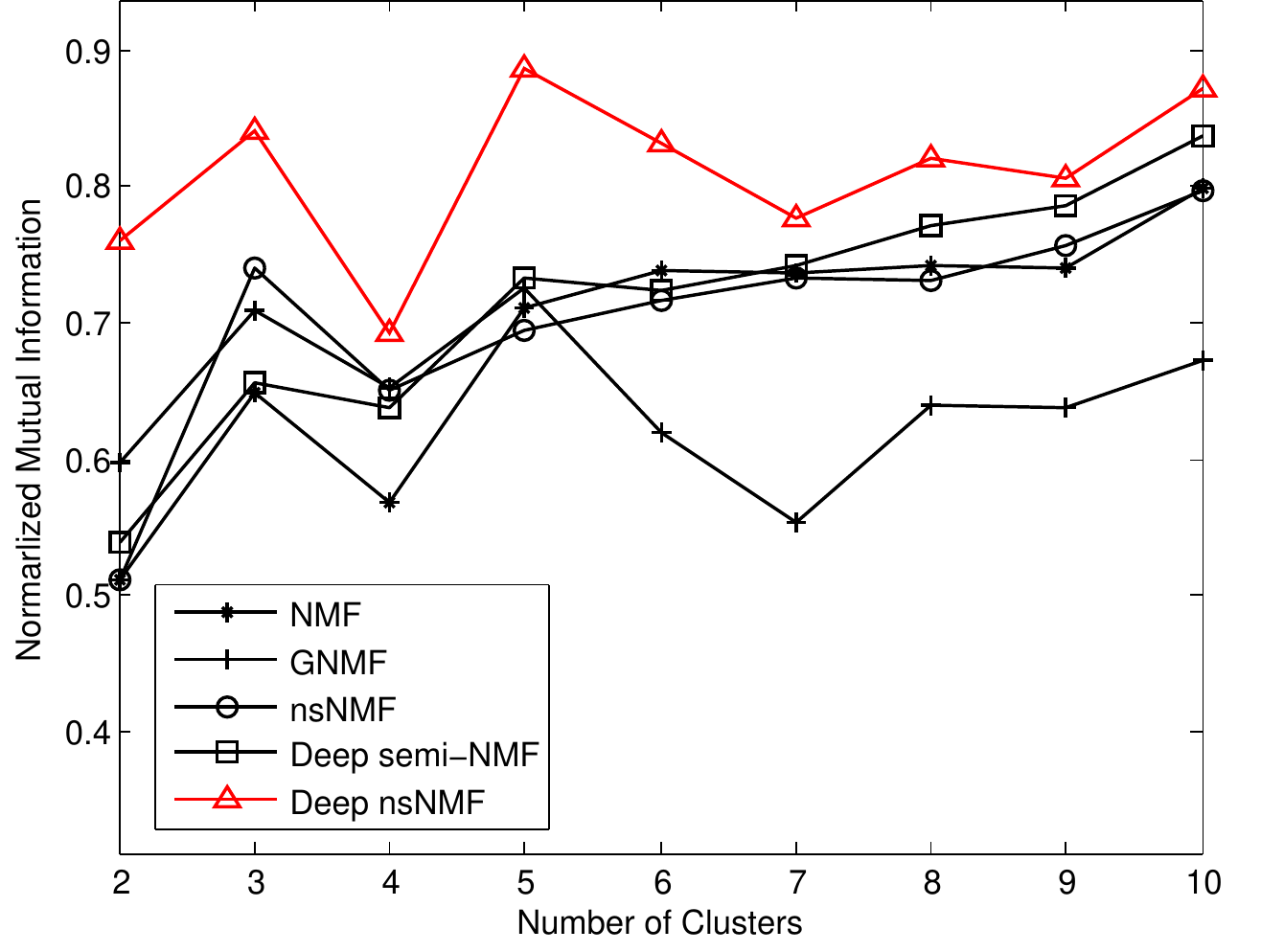}}
\caption{Clustering performance on the AT\&T ORL Database}
\label{fig_H_orl}
\end{figure*}

\begin{figure*}
\centering
\subfloat[Accuracy vs. Number of Clusters]{\includegraphics[width=0.4\textwidth]
{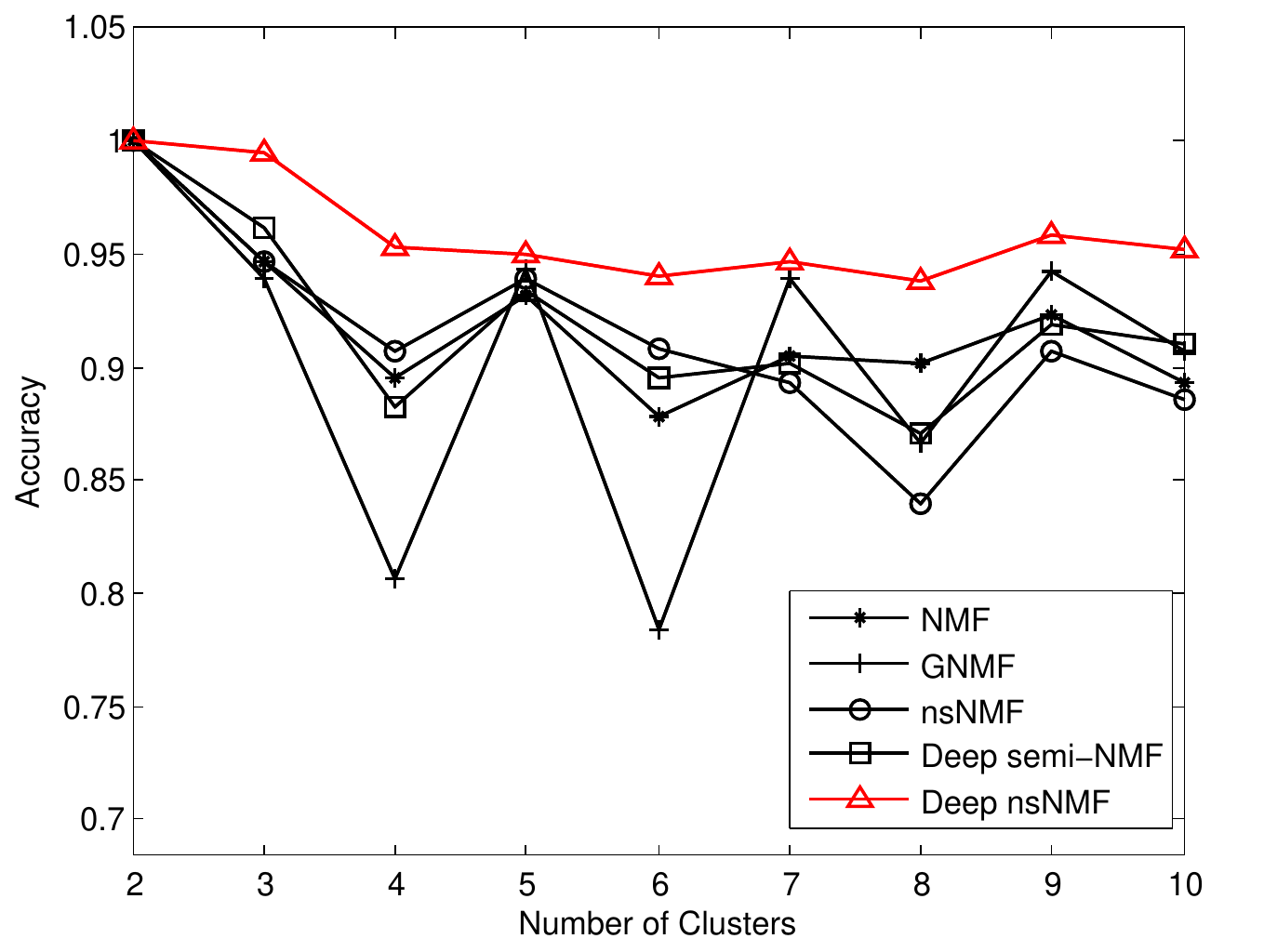}}
\subfloat[Mutual Information vs. Number of Clusters]{\includegraphics[width=0.4\textwidth]
{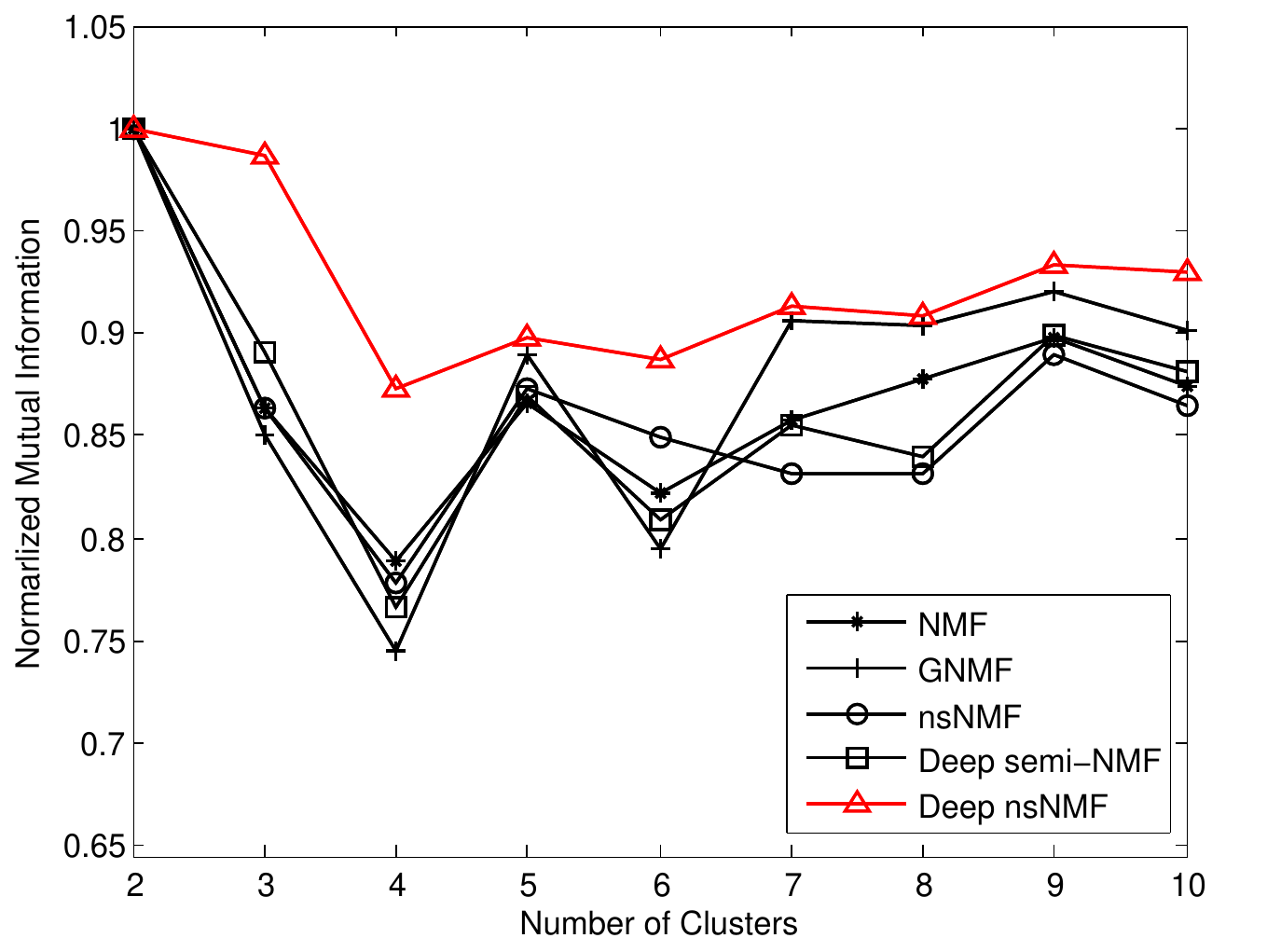}}
\caption{Clustering performance on the JAFFE Database}
\label{fig_H_jaffe}
\end{figure*}
\begin{figure*}
\centering
\subfloat[Accuracy vs. Number of Clusters]{\includegraphics[width=0.4\textwidth]
{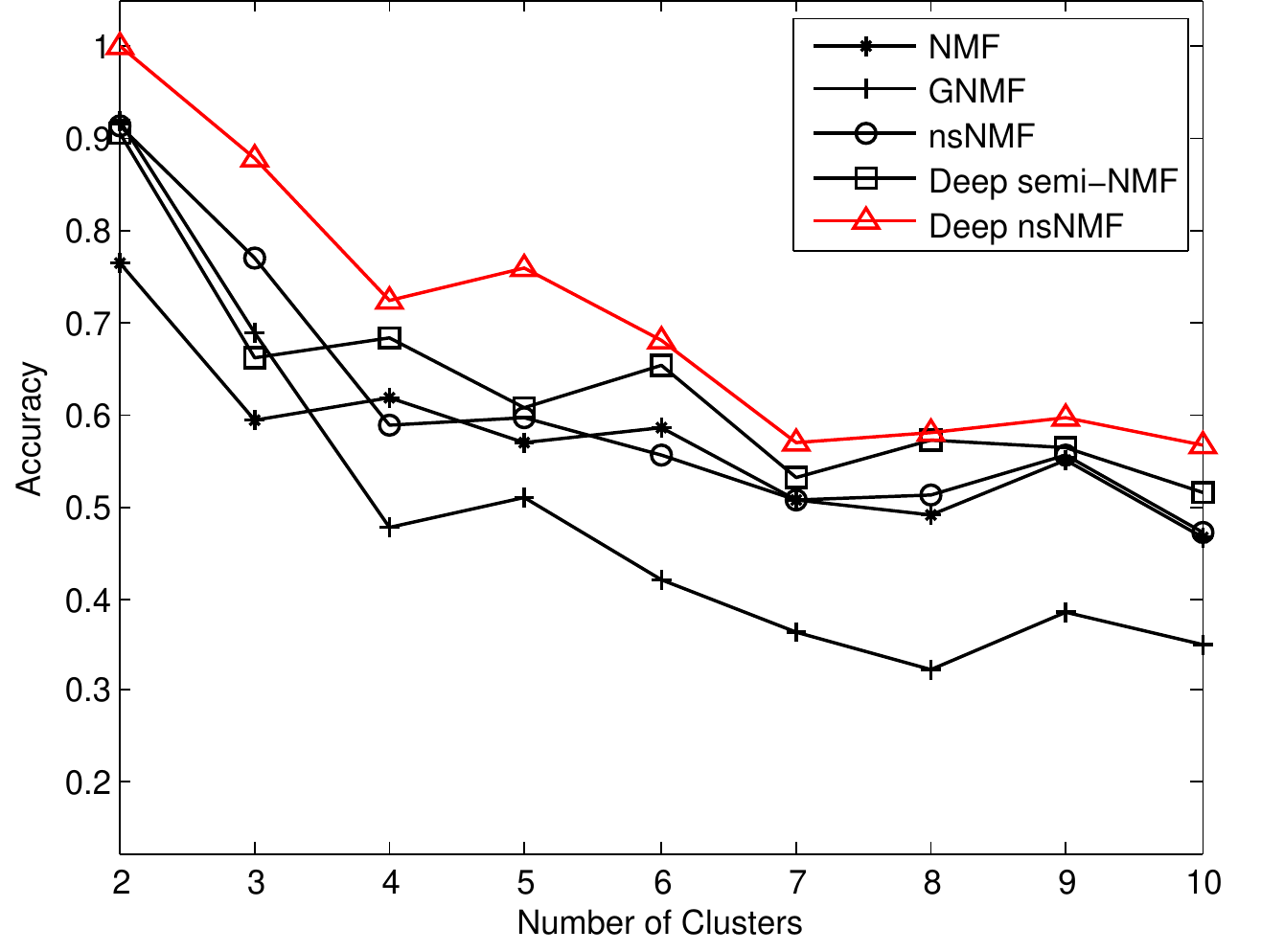}}
\subfloat[Mutual Information vs. Number of Clusters]{\includegraphics[width=0.4\textwidth]
{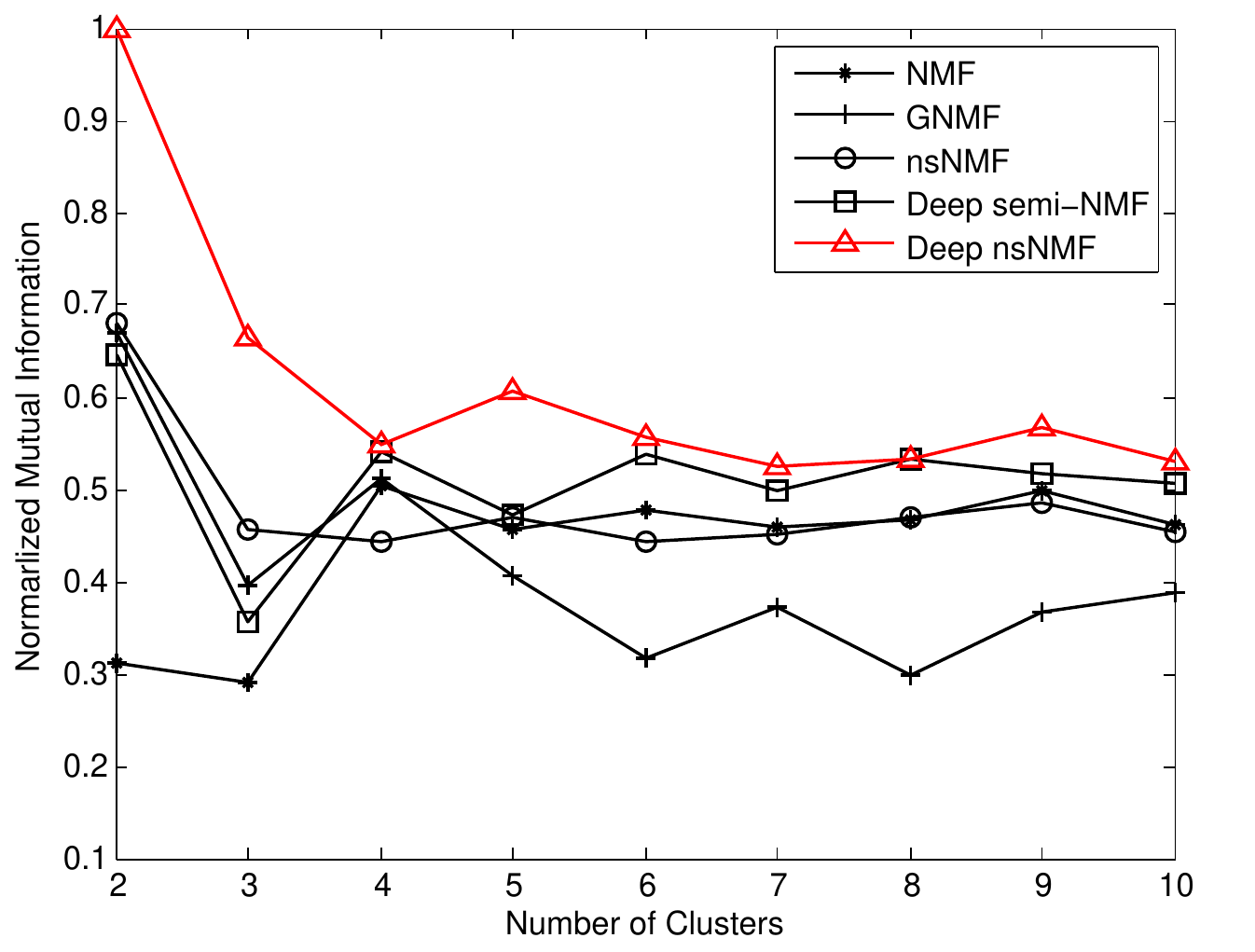}}
\caption{Clustering performance on the Yale Database}
\label{fig_H_yale}
\end{figure*}

\begin{table*}
\centering \caption{ Clustering Performance on AT\&T ORL Database } \label{table_orl}
\renewcommand{\arraystretch}{1.3}
\doublerulesep=1pt
\begin{tabular}{|c||c|c|c|c|c||c|c|c|c|c|}
\hline
 ~   & \multicolumn{5}{c||}{Accuracy (\%)} & \multicolumn{5}{c|}{Normalized Mutual Information (\%)}\\
 \cline{2-11}
 $K$ & NMF & GNMF & \emph{ns}NMF & Deep Semi-NMF & Deep \emph{ns}NMF & NMF & GNMF & \emph{ns}NMF & Deep Semi-NMF & Deep \emph{ns}NMF \\
 \hline
2& 85.00 & 89.00 & 85.00 & 86.00 & \textbf{95.00}
 & 51.05 & 59.73 & 51.05 & 53.85 & \textbf{76.10} \\
 \hline
3& 80.00 & 81.00 & 83.33 & 81.00 & \textbf{93.33}
 & 64.90 & 70.97 & 74.03 & 65.60 & \textbf{84.11} \\
 \hline
4& 69.25 & 77.15 & 73.50 & 72.25 & \textbf{80.00}
 & 56.72 & 65.26 & 65.10 & 63.78 & \textbf{69.25} \\
 \hline
5& 74.60 & 74.00 & 75.20 & 77.40 & \textbf{94.60}
 & 71.12 & 72.52 & 69.38 & 73.32 & \textbf{88.75} \\
 \hline
6& 67.17 & 52.50 & 67.67 & 70.00 & \textbf{81.67}
 & 73.81 & 61.97 & 71.73 & 72.34 & \textbf{83.17} \\
 \hline
7& 69.00 & 54.43 & 69.00 & 72.43 & \textbf{79.43}
 & 73.67 & 55.26 & 73.27 & 74.24 & \textbf{77.75} \\
 \hline
8& 74.13 & 52.13 & 74.50 & 77.63 & \textbf{82.13}
 & 74.29 & 63.98 & 73.20 & 77.18 & \textbf{82.02} \\
 \hline
9& 66.22 & 58.67 & 69.67 & 71.33 & \textbf{79.78}
 & 74.04 & 63.84 & 75.65 & 78.62 & \textbf{80.73} \\
 \hline
10& 70.70 & 57.70 & 69.00 & 77.10 & \textbf{78.50}
  & 79.82 & 67.21 & 79.71 & 83.83 & \textbf{87.33} \\
 \hline
Avg. & 72.90 & 66.29 & 74.10 & 76.13 & \textbf{84.94}
     & 68.82 & 64.53 & 70.35 & 71.42 & \textbf{81.02} \\
\hline
\end{tabular}
\end{table*}
\begin{table*}
\centering \caption{ Clustering Performance on JAFFE Database } \label{table_jaffe}
\renewcommand{\arraystretch}{1.3}
\doublerulesep=1pt
\begin{tabular}{|c||c|c|c|c|c||c|c|c|c|c|}
\hline
 ~   & \multicolumn{5}{c||}{Accuracy (\%)} & \multicolumn{5}{c|}{Normalized Mutual Information (\%)}\\
 \cline{2-11}
 $K$ & NMF & GNMF & \emph{ns}NMF & Deep Semi-NMF & Deep \emph{ns}NMF & NMF & GNMF & \emph{ns}NMF & Deep Semi-NMF & Deep \emph{ns}NMF \\
 \hline
2& \textbf{100.00} & \textbf{100.00} & \textbf{100.00} & \textbf{100.00} & \textbf{100.00}
  & \textbf{100.00} & \textbf{100.00} & \textbf{100.00} & \textbf{100.00} & \textbf{100.00} \\
 \hline
3& 94.68 & 93.87 & 94.68 & 96.13 & \textbf{99.52}
 & 86.32 & 85.01 & 86.32 & 89.13 & \textbf{98.71} \\
 \hline
4& 89.54 & 80.69 & 90.69 & 88.28 & \textbf{95.29}
 & 78.94 & 74.51 & 77.77 & 76.65 & \textbf{87.35} \\
 \hline
5& 93.14 & 94.38 & 93.90 & 93.43 & \textbf{94.95}
 & 86.59 & 88.90 & 87.35 & 86.89 & \textbf{89.76} \\
 \hline
6& 87.85 & 78.38 & 90.85 & 89.54 & \textbf{94.00}
 & 82.21 & 79.52 & 84.98 & 80.94 & \textbf{88.76} \\
 \hline
7& 90.54 & 93.89 & 89.26 & 90.20 & \textbf{94.70}
 & 85.81 & 90.58 & 83.10 & 85.50 & \textbf{91.29} \\
 \hline
8& 90.12 & 86.67 & 83.92 & 87.08 & \textbf{93.80}
 & 87.73 & 90.42 & 83.14 & 83.98 & \textbf{90.87} \\
 \hline
9& 92.36 & 94.24 & 90.73 & 91.88 & \textbf{95.81}
 & 89.75 & 92.05 & 88.93 & 89.87 & \textbf{93.36} \\
 \hline
10& 89.34 & 90.66 & 88.59 & 91.08 & \textbf{95.16}
  & 87.44 & 90.15 & 86.49 & 88.16 & \textbf{92.96} \\
 \hline
Avg. & 91.95 & 90.31 & 91.40 & 91.96 & \textbf{95.91}
     & 87.20 & 87.90 & 86.45 & 86.79 & \textbf{92.56} \\
\hline
\end{tabular}
\end{table*}
\begin{table*}
\centering \caption{ Clustering Performance on Yale Database } \label{table_yale}
\renewcommand{\arraystretch}{1.3}
\doublerulesep=1pt
\begin{tabular}{|c||c|c|c|c|c||c|c|c|c|c|}
\hline
 ~   & \multicolumn{5}{c||}{Accuracy (\%)} & \multicolumn{5}{c|}{Normalized Mutual Information (\%)}\\
 \cline{2-11}
 $K$ & NMF & GNMF & \emph{ns}NMF & Deep Semi-NMF & Deep \emph{ns}NMF & NMF & GNMF & \emph{ns}NMF & Deep Semi-NMF & Deep \emph{ns}NMF \\
 \hline
 2& 76.36 & 91.82 & 91.36 & 90.45 & \textbf{100.00}
  & 31.09 & 66.87 & 67.93 & 64.73 & \textbf{100.00} \\
 \hline
3& 59.39 & 68.79 & 76.97 & 66.06 & \textbf{87.88}
 & 29.21 & 39.59 & 45.75 & 35.59 & \textbf{66.46} \\
 \hline
4& 61.82 & 47.73 & 58.86 & 68.18 & \textbf{72.27}
 & 50.32 & 51.29 & 44.24 & 54.13 & \textbf{54.75} \\
 \hline
5& 56.91 & 51.09 & 59.64 & 60.73 & \textbf{76.00}
 & 45.76 & 40.76 & 47.06 & 47.14 & \textbf{60.61} \\
 \hline
6& 58.48 & 42.12 & 55.61 & 65.45 & \textbf{68.03}
 & 47.64 & 31.67 & 44.46 & 53.81 & \textbf{55.59} \\
 \hline
7& 50.65 & 36.36 & 50.65 & 53.12 & \textbf{57.01}
 & 45.83 & 37.13 & 45.19 & 49.75 & \textbf{52.60} \\
 \hline
8& 49.20 & 32.39 & 51.36 & 57.27 & \textbf{58.07}
 & 46.66 & 29.81 & 46.88 & 53.34 & \textbf{53.23} \\
 \hline
9& 55.15 & 38.59 & 55.56 & 56.36 & \textbf{59.60}
 & 49.76 & 36.58 & 48.66 & 51.66 & \textbf{56.73} \\
 \hline
10& 46.73 & 35.09 & 47.09 & 51.64 & \textbf{56.64}
  & 46.29 & 38.92 & 45.27 & 50.58 & \textbf{52.99} \\
 \hline
Avg. & 57.19 & 49.33 & 60.79 & 63.25 & \textbf{70.61}
     & 43.62 & 41.40 & 48.38 & 51.19 & \textbf{61.44} \\
\hline
\end{tabular}
\end{table*}

\figurename \ref{fig_H_orl}-\ref{fig_H_yale} plot the averaged Accuracy and normalized mutual information against different number of clusters on the AT\&T ORL, JAFFE and Yale databases, respectively, and \tablename{} \ref{table_orl}-\ref{table_yale} list the detailed clustering accuracy and normalized mutual information. It can be observed that:
%
%
\begin{enumerate}
\item Compared with NMF, \emph{ns}NMF, and GNMF, which are developed to provide part-based representations but without deep architectures, the dnsNMF algorithm always achieved the best clustering performance. This may suggest that a deep architecture does help to improve the performance of NMF.
\item Compared to Deep semi-NMF, which has a deep architecture but without parts-based representation guarantee, the dnsNMF achieved improved accuracy. This suggests that, by leveraging the power of both \emph{ns}NMF and deep architectures, the dnsNMF can learn much better features representation for the data than other methods.
\end{enumerate}

\subsection{Effect of Depth of  dnsNMF}
\begin{figure}
\centering
\includegraphics[width=0.4\textwidth]
{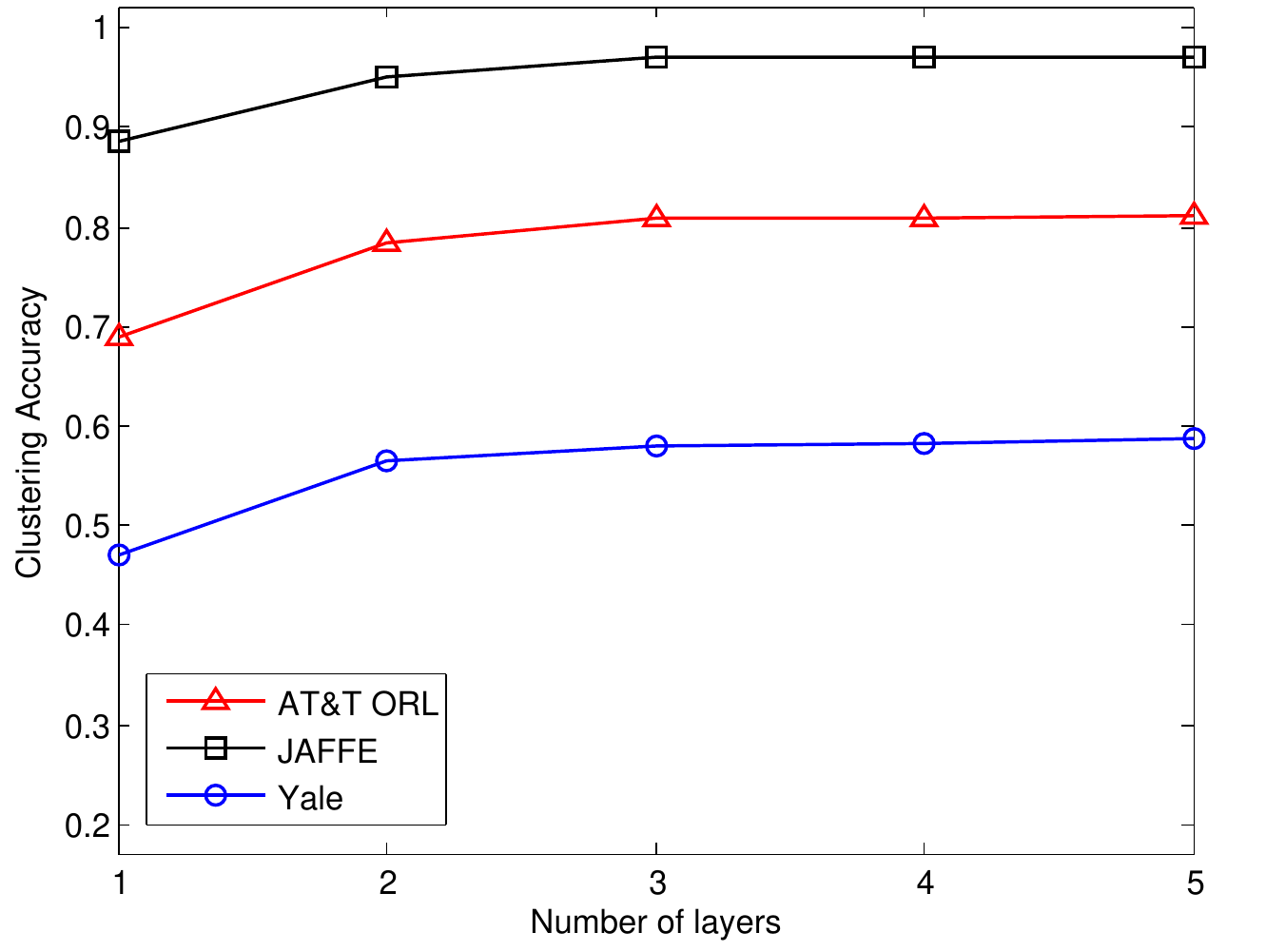}
\caption{Effect of depth of the dnsNMF on AT\&T ORL, JAFFE and Yale data sets. It illustrates the clustering accuracy of the dnsNMF with depth varying from 1 to 5. It is easy to note that in these three data sets the clustering accuracy consistently increases with increasing depth at the early stage and then is nearly invariable with the added layers.}
\label{depth}
\end{figure}

In this section, we study how the depth (number of layers) of dnsNMF affects the clustering accuracy. We randomly chose 10 categories from the data set and shuffled the images from these 10 categories for clustering analysis. Then, we applied the dnsNMF method with depth varying from 1 layer to 5 layers to reconstruct \mat{X}. In all the cases, the dimensionality of the toppest layer was fixed to be ten; and the other parameters were chosen using grid search to achieve the best results. The algorithm was tested on AT\&T ORL, JAFFE and Yale data sets, respectively, and the results averaged over 10 random runs were plotted in \figurename \ref{depth}. From the figure, the clustering accuracy consistently increases with increasing depth at the early stage and then is nearly invariable with the depth. In these three data sets, it seems that three layers are sufficient to produce satisfactory accuracy. One may question why the performance does keep increasing when the depth further increases. We guess it is because that it is more and more difficult to choose optimal parameters when the number of depth increase gradually. Moreover, the depth efficiency may also rely on the scale of data at hand. Hence, how to choose a proper depth in practical applications deserves further study.

\section{Conclusions and Future Works}
In this paper, we propose a deep nsNMF method which equipes \emph{ns}NMF with a deep architecture. The new method is efficient to learn not only part-based but also hierarchical features. We also showed that how dnsNMF can be viewed as a restricted deep autoencoder. Experiments on AT\&T ORL, JAFFE and Yale data sets showed that the dnsNMF method can substantially improve clustering accuracy by moderately increasing the depth, demonstrating great potential of dnsNMF as a un-supervising learning method.

Some questions remain to be studied in our future work:
\begin{enumerate}
  \item How to choose a proper depth in practice to achieve a good tradeoff between the accuracy and the efficiency. While we believe that the scale of data affect the depth, a further study on this topic is needed.

  \item Although we have shown that dnsNMF is a restricted deep autoencoder, comparison with deep autoencoder with different activation functions is missing. A comprehensive comparative study  could be very interesting.
\end{enumerate}

\appendices

\section{Proof of Proposition 1}\label{appendix A}
\noindent \textbf{Proof.} According to \eqref{eq:dnsNMFobj_zi}, the gradient of $E_{Deep}{\left({\bf Z}_{i}\right)}$
\begin{equation}
\triangledown_{{\bf Z}_{i}}E_{deep}({\bf Z}_{i})={\bf A}_{i}^{T}{\bf A}_{i}{\bf Z}_{i}{\bf B}_{i}{\bf B}_{i}^{T}-{\bf A}_{i}^{T}{\bf X}{\bf B}_{i}^{T}.
\end{equation}

For any two matrices ${\bf Z}_{i1}, {\bf Z}_{i2}\in{\bf\Re}^{r_{i-1}\times r_{i}}$, we have
\begin{eqnarray}\label{eq:proof_of_prop1}
&&\|\triangledown_{{\bf Z}_{i}}E_{deep}({\bf Z}_{i1})-\triangledown_{{\bf Z}_{i}}E_{deep}({\bf Z}_{i2})\|_{F}^{2}\nonumber
\\&&=\|{\bf A}_{i}^{T}{\bf A}_{i}({\bf Z}_{i1}-{\bf Z}_{i2}){\bf B}_{i}{\bf B}_{i}^{T}\|_{F}^{2}\nonumber
\\&&=tr((\mats[a]{U}\mats[a]{\Sigma}\mats[a]{U}^{T}({\bf Z}_{i1}-{\bf Z}_{i2})\mats[b]{U}\mats[b]{\Sigma}\mats[b]{U}^{T}  )^{T}\nonumber
\\&&\qquad\quad(\mats[a]{U}\mats[a]{\Sigma}\mats[a]{U}^{T}({\bf Z}_{i1}-{\bf Z}_{i2})\mats[b]{U}\mats[b]{\Sigma}\mats[b]{U}^{T})),
\end{eqnarray}
where $\mats[a]{U}\mats[a]{\Sigma}\mats[a]{U}^{T}$ and $\mats[b]{U}\mats[b]{\Sigma}\mats[b]{U}^{T}$ are the SVD decomposition of ${\bf A}_{i}^{T}{\bf A}_{i}$ and ${\bf B}_{i}{\bf B}_{i}^{T}$ respectively. and their singular values are $\{\delta_{1},\dots,\delta_{r_{i-1}} \} $ and $\{\sigma_{1},\dots,\sigma_{r_{i}} \}$, which are arranged in descending order. By simple algebraic multiplication, \eqref{eq:proof_of_prop1} is equivalent to
\begin{eqnarray}\label{eq:proof1_of_prop1}
&&\|\triangledown_{{\bf Z}_{i}}E_{deep}({\bf Z}_{i1})-\triangledown_{{\bf Z}_{i}}E_{deep}({\bf Z}_{i2})\|_{F}^{2}\nonumber
\\&&=tr(\mats[a]{\Sigma}^{2}\mats[a]{U}^{T}({\bf Z}_{i1}-{\bf Z}_{i2})\mats[b]{U}\mats[b]{\Sigma}^{2}\mats[b]{U}^{T}({\bf Z}_{i1}-{\bf Z}_{i2})^{T}\mats[a]{U})\nonumber
\\&&\le\delta_{1}^{2}tr(\mats[a]{U}^{T}({\bf Z}_{i1}-{\bf Z}_{i2})\mats[b]{U}\mats[b]{\Sigma}^{2}\mats[b]{U}^{T}({\bf Z}_{i1}-{\bf Z}_{i2})^{T}\mats[a]{U})\nonumber
\\&&=\delta_{1}^{2}tr(\mats[b]{\Sigma}^{2}\mats[b]{U}^{T}({\bf Z}_{i1}-{\bf Z}_{i2})^{T}({\bf Z}_{i1}-{\bf Z}_{i2})\mats[b]{U})\nonumber
\\&&\le\delta_{1}^{2}\sigma_{1}^{2}tr(\mats[b]{U}^{T}({\bf Z}_{i1}-{\bf Z}_{i2})^{T}({\bf Z}_{i1}-{\bf Z}_{i2})\mats[b]{U})\nonumber
\\&&=\delta_{1}^{2}\sigma_{1}^{2}tr(({\bf Z}_{i1}-{\bf Z}_{i2})^{T}({\bf Z}_{i1}-{\bf Z}_{i2}))\nonumber
\\&&=\delta_{1}^{2}\sigma_{1}^{2}\|{\bf Z}_{i1}-{\bf Z}_{i2}\|_{F}^{2}.
\end{eqnarray}
where $\delta_{1}$ and $\sigma_{1}$ are the largest singular value of $ {\bf A}_{i}^{T}{\bf A}_{i}$ and ${\bf B}_{i}{\bf B}_{i}^{T} $, respectively. Note that, the above equations come from the fact that $\mats[a]{U}\trans{\mats[a]{U}}$ and $\mats[b]{U}\trans{\mats[b]{U}}$ are identity matrices. From the \eqref{eq:proof1_of_prop1}, we have
\begin{eqnarray}
&&\|\triangledown_{{\bf Z}_{i}}E_{deep}({\bf A}_{i},{\bf Z}_{i1},{\bf B}_{i})-\triangledown_{{\bf Z}_{i}}E_{deep}({\bf A}_{i},{\bf Z}_{i2},{\bf B}_{i})\|_{F}\nonumber
\\&&\le L_{i}\|{\bf Z}_{i1}-{\bf Z}_{i2}\|_{F}.
\end{eqnarray}
Therefore, $\triangledown_{{\bf Z}_{i}}E_{deep}({\bf A}_{i},{\bf Z}_{i},{\bf B}_{i})$ is Lipschitz continuous and the Lipschitz constant is the product of the largest values of $ {\bf A}_{i}^{T}{\bf A}_{i}$ and ${\bf B}_{i}{\bf B}_{i}^{T} $, i.e., $L_{i}=\|{\bf A}_{i}^{T}{\bf A}_{i}\|_{2}*\|{\bf B}_{i}{\bf B}_{i}^{T}\|_{2}$. This completes the proof.

\ifCLASSOPTIONcaptionsoff
  \newpage
\fi

\bibliographystyle{IEEEtran}
\bibliography{yjsbib}

\end{document}